\documentclass[sigplan,10pt]{acmart}
\copyrightyear{2026}
\acmYear{2026}
\setcopyright{cc}
\setcctype{by}
\acmConference[Extended Version - EUROSYS '26]{21st European Conference on Computer Systems}{April 27--30, 2026}{Edinburgh, Scotland Uk}
\acmBooktitle{21st European Conference on Computer Systems (EUROSYS '26), April 27--30, 2026, Edinburgh, Scotland Uk}
\acmDOI{10.1145/3767295.3803595}
\acmISBN{979-8-4007-2212-7/2026/04}

\usepackage{algorithmic}
\usepackage{graphicx}
\usepackage{textcomp}
\usepackage{xcolor}
\usepackage{hyperref}
\hypersetup{colorlinks=true,urlcolor=cyan}

\usepackage[normalem]{ulem}
\usepackage[utf8]{inputenc}
\usepackage[T1]{fontenc}
\usepackage{paralist}

\usepackage{tikz}
\usepackage{comment}
\usepackage{multirow}
\usepackage{array}
\usepackage{longtable}
\usepackage[font=small,format=plain,labelfont=bf,textfont=bf]{caption}
\usepackage{subcaption}
\usepackage{algorithmic}
\usepackage{graphicx}
\usepackage{soul}
\usepackage{balance}
\usepackage{tabularx}
\usepackage{mdframed}   % for framing
\usepackage[frozencache,cachedir=minted-cache]{minted}
\usepackage{float}
\usepackage{booktabs}
\usepackage{listings}
\newcommand{\smartparagraph}[1]{\noindent{\bf #1.}\ }

\setminted{
linenos=true,
autogobble,
}

\newenvironment{code}
{\minted[escapeinside=@@,xleftmargin=15pt,fontsize=\footnotesize,baselinestretch=0.9]{python}}
{\endminted}
\newcommand*\circled[1]{\tikz[baseline=(char.base)]{
            \node[shape=circle,draw,inner sep=1pt] (char) {#1};}}

\newcommand{\ignore}[1]{}
\newcommand{\mask}{\textit{Mask}}
\newcommand{\bitval}{\textit{Bitval}}

\ifdefined\COMMENTS
    \newcommand{\rtodo}[1]{\textcolor{red}{\textbf{#1}}}
    \newcommand{\todo}[1]{\textcolor{red}{\textbf{TODO: #1}}}

    \newcommand{\remove}[1]{{\textcolor{red}{\st{#1}}}}
    \newcommand{\rremove}[2]{{\textcolor{red}{\st{#1} \textbf{(#2)}}}}
\else
    \newcommand{\rtodo}[1]{}
    \newcommand{\todo}[1]{}

    \newcommand{\remove}[1]{}
    \newcommand{\rremove}[2]{}
\fi

\begin{document}

\title{
Reducing the GPU Memory Bottleneck with Lossless Compression for ML
}

\author{Aditya K Kamath}
\affiliation{%
  \institution{University of Washington}
  \city{Seattle}
  \state{WA}
  \country{USA}
}
\email{akkamath@cs.washington.edu}

\author{Arvind Krishnamurthy}
\affiliation{%
  \institution{University of Washington}
  \city{Seattle}
  \state{WA}
  \country{USA}
}
\email{arvind@cs.washington.edu}

\author{Marco Canini}
\affiliation{%
  \institution{KAUST}
  \city{Thuwal}
  \country{Saudi Arabia}
}
\email{marco@kaust.edu.sa}

\author{Simon Peter}
\affiliation{%
  \institution{University of Washington}
  \city{Seattle}
  \state{WA}
  \country{USA}
}
\email{simpeter@cs.washington.edu}

\begin{abstract}
Machine learning (ML) training and inference often process data sets far exceeding GPU memory capacity,
forcing them to rely on PCIe for on-demand tensor transfers, causing critical transfer bottlenecks. 
Lossy compression has been proposed to relieve bottlenecks but introduces workload-dependent accuracy loss, making it complex or even prohibitive to use in existing ML deployments.

We explore lossless compression as an alternative that avoids this deployment complexity.
We identify where lossless compression can be integrated into ML pipelines while minimizing interference with GPU execution.
Based on our findings, we introduce Invariant Bit Packing (IBP), a novel lossless compression algorithm designed to minimize data transfer time for ML.
IBP identifies and eliminates invariant bits across groups of tensors, improving throughput through GPU-optimized decompression that leverages warp parallelism, low-overhead bit operations, and asynchronous PCIe transfers. We provide easy-to-use APIs, showcasing them by adding IBP support to GNN training, as well as DLRM and LLM inference frameworks. 
IBP achieves, on average, 74\% faster GNN training, 180\% faster DLRM embedding lookup, and 24\% faster LLM inference.

\end{abstract}



\maketitle

% Use the following at camera-ready time to suppress page numbers.
% Comment it out when you first submit the paper for review.

%\thispagestyle{empty}

\section{Introduction}

Machine learning (ML) models, such as Graph Neural Networks (GNNs), Deep Learning Recommendation Models (DLRMs), and Large Language Models (LLMs) have found widespread use across various domains, such as e-commerce, knowledge graph processing, drug discovery, and fraud detection. Training and inference with these models allow for efficient learning on semi-structured data~\cite{GNN_recommendation:SIGKDD:2018, KG_GNN:Neurips:2021, pick_choose:WWW:2021, Drug_GNN:Neurocomputing:2021} by processing tensors (i.e., vertex features for GNNs, embedding entries for DLRMs, KV-cache/weight tensors for LLMs) on the scale of hundreds of gigabytes to terabytes, exceeding the memory capacity of available GPUs. To handle this scale, CPU memory~\cite{Legion:ATC:2023, flexgen:ICML:2023}, disk~\cite{mariusgnn:Eurosys:2023}, and networked storage servers~\cite{BGL:NSDI:2023} are used, which are orders of magnitude larger than GPU memory. Tensors are then transferred on-demand to the GPU.

On-demand data loading causes execution to be bottlenecked by PCIe bandwidth~\cite{silod:eurosys:2023, infinigen:OSDI:2024}. 
A significant portion of time is spent waiting for the required data to reach the GPU~\cite{ducati:AMD:2023}.
Caching~\cite{Pagraph:SoCC:2020, BGL:NSDI:2023, ducati:AMD:2023, Legion:ATC:2023, ugache:SOSP:2023,  gnnlab:Eurosys:2022} and overlapping compute and data transfer provide some benefit, but PCIe bandwidth is over an order of magnitude lower than GPU memory bandwidth~\cite{v100, a100, h100}. Even 10\% of cache misses can double the fetch overhead, and if designed inefficiently, PCIe memory fetching can slow down workloads manyfold~\cite{feature_comp:VLDB:2024}.
Proposals to address this issue typically involve lossy compression. 
For example, quantization~\cite{quantization, feature_comp:VLDB:2024, bifeat:arxiv:2023} is a popular compaction method reducing memory footprint and data transfers.
Unfortunately, lossy compression degrades model quality to an extent that is difficult to predict and varies from model to model~\cite{dutta2024accuracyneed}.
This trade-off is undesirable for commercial applications, due to the risk of revenue loss~\cite{bagpipe:SOSP:2023, dlrm_efficiency:HPCA:2021}.

We investigate lossless compression as an alternative without the deployment complexity.
Making GPU-based lossless compression practical for ML data transfers is a difficult task.
CPU-GPU transfer latencies are competitive with GPU decompression latency, making traditional GPU-accelerated compression algorithms heavyweight, and reliance on the host CPU in the critical path can eliminate efficiency gains.
In addition, extraneous GPU-resident decompression data structures compete with the model and the input's GPU memory usage, making the use of large buffers for decompression impractical. For example, in GNNs the input tensors of a single batch are on the order of gigabytes. Reducing GPU memory usage of input requires the batch size to be reduced, necessitating more batches per training epoch~\cite{10.14778/3717755.3717776}. Similarly for LLMs, the GPU memory size limits the context length and batch sizes of requests during inference~\cite{PagedAttention:SOSP:2023}.

We address these challenges by (1) choosing a data layout that enables efficient PCIe transfers directly from the GPU, (2) performing data compression in CPU memory while minimizing GPU-side metadata, and (3) optimizing GPU decompression for just-in-time use.
Unlike lossy compression, we avoid any information loss trade-off.

Motivated by these insights, we introduce Invariant Bit Packing (IBP), a GPU-optimized lossless compression algorithm tailored to ML training and inference acceleration. IBP analyzes input data to identify invariant bits shared across tensors. These bits are stored in a fixed-size metadata region (on the order of KBs) within GPU memory and removed from the compressed data items. We provide a parallel implementation of IBP that enables GPU decompression faster than a CPU-to-GPU transfer. IBP utilizes asynchronous, aligned memory accesses~\cite{bamasplos}, low-overhead warp primitives~\cite{cuda_warp_prim}, and cheap bitshift operations, achieving low decompression overhead and high throughput (\autoref{tab:comp_perf}). 
IBP makes use of zero-copy~\cite{cuda_uvm} to directly fetch and decompress CPU memory from the GPU.
The principles we introduce could also apply to disk and network transfers.

Our contributions are as follows:
\begin{itemize}
    \item We analyze PCIe transfer bottlenecks in GNN, DLRM, and LLM workloads, examine existing lossy and lossless compression, and highlight why prior approaches fail to deliver performance in ML pipelines (\S\ref{sec:background}).
    \item We introduce Invariant Bit Packing (IBP), an optimized lossless compression algorithm that removes invariant bits across tensors with diverse datatypes (e.g., float32, float16, bfloat16), using minimal metadata (\S\ref{sec:motivation}--\ref{sec:ibp}). Warp-parallel decompression and optimized PCIe data transfer achieve low decompression overhead and high throughput.
    \item We extend popular ML systems by incorporating IBP into GPU software caches and PCIe traffic paths (\S\ref{sec:ibp_for_ml}). Our PyTorch extension and CUDA library simplify adoption across training and inference workloads.
    \item We evaluate IBP against state-of-the-art GPU-accelerated compression libraries (nvCOMP~\cite{nvcomp}, ndzip-gpu~\cite{knorr:ndzip:2021}) on real datasets (\S\ref{sec:eval}). IBP accelerates GNN training by 74\%, DLRM embedding lookup by 180\%, and LLM inference by 24\% on average on an A100 GPU, while preserving model accuracy. We demonstrate that IBP works on streaming datasets via sampling (\S\ref{subsec:sampling}).
\end{itemize}
We have made our code available at \url{https://github.com/AKKamath/InvariantBitPacking} to aid future research.

\section{Background} \label{sec:background}

Modern ML models increasingly operate on data that significantly exceeds the capacity of available GPU memory. They depend on transferring data from CPU memory over PCIe, introducing critical bottlenecks where PCIe bandwidth (rather than GPU compute) becomes the performance constraint.
Latency-hiding techniques like prefetching help hide the transfer latency by fetching data for the next batch while compute for the current batch is ongoing. However, if the transfer latency far exceeds the compute latency, the transfer remains the bottleneck. For example, we evaluate GNN training where the training computation and CPU-GPU transfers are done in parallel, yet training is frequently waiting for the data to arrive (\autoref{sec:gnn_speedup}).

We analyze the origin and impact of PCIe transfers in three representative ML workloads---GNN training and DLRM and LLM inference---illustrating why such transfers are a dominant performance constraint (\S\ref{sec:pcie_in_ml}). We then examine existing techniques to mitigate this overhead, focusing on both lossy and lossless compression, and discuss the trade-offs each approach introduces in ML (\S\ref{sec:compression_ml}).

\subsection{PCIe Transfers in Training and Inference}\label{sec:pcie_in_ml}

\smartparagraph{GNN training} GNNs are machine learning models that learn properties of graph-based data.
For classification problems, GNNs use three types of inputs: a graph topology, a set of feature vectors, and a set of labels.
The graph topology contains relations between nodes and edges of the graph in a sparse format.
The feature vectors add more information to the graph, associated with the nodes or edges.
The feature vectors can be dense floating point vectors~\cite{ogb:Neurips:2020} or sparse bitmask vectors~\cite{citeseer_dataset}.
The labels then group the nodes or edges of the graph.
A concrete example is a citation graph, where nodes are published papers and edges are citations. Each node in the graph has a feature vector keeping track of certain keywords used in the paper.
The labels finally classify each paper by its field of research.
A GNN uses this information to predict labels for papers that lack them.

Modern graph datasets exceed the capacity of a GPU, and so \textit{graph sampling} is used to construct minibatches for GNN training~\cite{gnnsampling}.
For each minibatch: \circled{1} a breadth-first-search (BFS) is performed, constructing a subgraph whose nodes' feature vectors are \circled{2} fetched to the GPU from the CPU for \circled{3} GNN model training.
These steps are repeated until training is complete.
The fetching phase \circled{2}, where data is transferred across PCIe, is the primary focus of this work. 

GNNs already make use of a static cache initialized during preprocessing to reduce data movement~\cite{ugache:SOSP:2023, ducati:AMD:2023, Legion:ATC:2023, Pagraph:SoCC:2020, gnnlab:Eurosys:2022}. The cache has two main components: a data store and a hashmap. 
All the cached data together make up the data store, possibly discontiguous or distributed across multiple interconnected GPUs. 
The hashmap maps a given ID to the relevant cached item.
For GNNs, the node/edge ID is passed to the hashmap, which returns a pointer to the feature vector in the data store associated with the ID. 
Since the cache cannot store all the required data, determining what to cache remains a key challenge~\cite{ugache:SOSP:2023, ducati:AMD:2023, feature_comp:VLDB:2024, Pagraph:SoCC:2020}.
Modern GNN frameworks~\cite{Legion:ATC:2023, ducati:AMD:2023} use preprocessing to choose which feature vectors to cache in GPU memory.

\smartparagraph{DLRM inference}
DLRMs are designed to infer facts from sparse categorical data like personal information.
The model takes two types of inputs: dense contiguous features (e.g., age) and sparse categorical features (e.g., hobbies).
The sparse features are transformed into dense embeddings and combined with the dense features for inference.

Each sparse feature acts as a hash key, used to look up an \textit{embedding table}.
These tables contain entries for each possible sparse feature, possibly extending to millions of rows, exceeding GPU memory capacity.
To handle the capacity requirements, tables can reside in CPU memory, with lookups moving data to the GPU~\cite{ugache:SOSP:2023}.
Each minibatch looks up thousands of table entries---bottlenecking DLRM inference~\cite{HugeCTR:RecSy:2022}. During inference, embedding tables are static.

\smartparagraph{LLM offloading}
LLMs operate on textual data converted to the form of floating-point tokens (typically 16-bit float, i.e., float16/bfloat16).
Each LLM layer uses Attention~\cite{Attention:Neurips:2017}, where each input token is multiplied by a query, key, and value weight to produce a query (Q), key (K), and value (V) tensor.
Every token's Q is operated on by the other tokens' K and V to determine their relative impact on the current token.
Every generated output token repeats this process, being operated on by the K and V of previous tokens.

KV-caching~\cite{PagedAttention:SOSP:2023} is a technique used by LLMs that avoids recomputing the K and V for each iteration by storing them the first time they are computed, allowing for re-use across output tokens. 
For modern LLMs reaching context lengths of millions of tokens, the size of the KV-cache far exceeds the size of GPU memory.
Offloading~\cite{flexgen:ICML:2023} is used to increase context lengths by storing the KV-cache in CPU memory, then transferring as needed to the GPU.
As the PCIe interconnect is bandwidth-limited, KV-cache offloading has become the critical bottleneck in large-context LLM inference~\cite{infinigen:OSDI:2024, flexgen:ICML:2023}.
Sparse KV-caching~\cite{infinigen:OSDI:2024, h2o:neurips:2023} brings only the critical KV subset onto the GPU, mitigating the bottleneck but not eliminating it, since KV transfers remain on the critical path.

The LLM models themselves are quite large, on the order of gigabytes, possibly exceeding the capacity of a GPU.
These weights are static during LLM inference.
Offloading has been used to keep the LLM weights in CPU memory, transferring them to the GPU layer-by-layer during inference~\cite{flexgen:ICML:2023}.
This allows a single GPU to serve much larger models, reducing the hardware cost of deployment. It also allows smaller devices to serve larger models.

\subsection{ML Compression Deployment Challenges} \label{sec:compression_ml}

Data compression can relieve these transfer bottlenecks, but is challenging to deploy in ML pipelines. Lossy compression reduces data sizes and often even omits an expensive decompression step. However, it loses information. It must therefore be applied judiciously to data that can tolerate loss in fidelity without compromising application utility. Lossless compression preserves the original data, allowing it to be deployed as a black box. However, lossless decompression has overhead, complicating its adoption in performance-critical sections of an application. In this section, we describe the use of both in of ML applications.
Throughout the paper, we use two terms to describe compression performance: $$\mbox{Space savings (\%)} = \left(1 - \frac{\mbox{Compressed Size}}{\mbox{Uncompressed Size}}\right) \times 100,$$
$$\mbox{Compression ratio} = \frac{\mbox{Uncompressed Size}}{\mbox{Compressed Size}}.$$

\smartparagraph{Lossy compression risks model accuracy loss}
In machine learning, quantization~\cite{quantization} has become the major method for lossy compression.
In quantization, statistical analysis is used to reduce the number of possible states that individual data elements take.
For example, 16-bit to 8-bit quantization allows data to be stored in half the capacity but reduces the number of representable states by a factor of 256.
As bits are reduced, information is lost.

Employing lossy compression in machine learning comes with risks and drawbacks.
As information is lost, unexpected model behavior can arise~\cite{dutta2024accuracyneed} and
model accuracy becomes compromised. Additionally, lossy compression requires careful tuning to specific model behaviors. Choices have to be made about which data to store in compressed form and how to manage outlier data to retain model accuracy~\cite{gptint8:neurips:2022}, which vary across machine learning models~\cite{qlora:Neurips:2023, feature_comp:VLDB:2024,atom:mlsys:2024}. Both factors greatly discourage the use of lossy compression in commercial models (e.g., DLRM), where even a 0.1\% loss in accuracy is considered unacceptable~\cite{dlrm_efficiency:HPCA:2021, bagpipe:SOSP:2023}.

\smartparagraph{Lossless compression has high decompression overhead}
Lossless compression sidesteps these risks by guaranteeing that data can be decompressed into its original form without any loss of information.
It simplifies adoption, as users do not need to worry about impacts on model accuracy, and they can treat compression as a black box for deployment. However, the high decompression overhead of lossless compression algorithms can lead to increases in latency and reduced training performance. For ML applications running on GPUs, these overheads can become pronounced.

Common code-based lossless algorithms, like Huffman Coding~\cite{huffman_code}, employ decompression code tables or dictionaries that scale with input sizes to obtain high compression ratios.
Input elements are mapped to bit patterns based on the frequency of element values, with more frequent values occupying fewer bits, effectively reducing data sizes.
Decompression involves reading a unit of data from the input buffer and then reading the relevant code table to map it to the output. 
It leads to multiple PCIe reads on the critical path for each element, degrading throughput. 
Further, large intermediate decompression buffers are infeasible as model inputs consume a significant fraction of GPU memory. Such compression algorithms are unsuitable for ML applications.

\begin{table*}[t]
\centering
\caption{Average speedup of compressed CPU-to-GPU transfers (best in \textbf{bold}).}
\label{tab:comp_perf}
\begin{tabular}{|c||c|c||c||c|c|c|}  \hline
\multirow{2}{*}{\textbf{Algorithm}} & \multirow{2}{*}{\textbf{\shortstack{GNN \\ Sparse}}} & \multirow{2}{*}{\textbf{\shortstack{GNN \\ Dense}}}  & \multirow{2}{*}{\textbf{\shortstack{DLRM \\Weights}}} & \multirow{2}{*}{\textbf{\shortstack{LLM \\ KV (FP16)}}}  & \multirow{2}{*}{\textbf{\shortstack{LLM \\ KV (BF16)}}} & \multirow{2}{*}{\textbf{\shortstack{LLM \\ Weights (BF16)}}}\\ 
 & & & & & & \\\hline
ANS & 2.1 & 0.1 & 0.1 & 0.10 & 0.12 & 0.20 \\\hline
Bitcomp & 3.9 & 0.4 & 0.5 & 0.65 & 0.65 & 0.64 \\\hline
Cascaded & 8.6 & 0.8 & 0.8 & 0.48 & 0.51 & 0.76 \\\hline
GDeflate & 2.1 & 0.3 & 0.5 & 0.16 & 0.16 & 0.18 \\\hline
LZ4 & 2.3 & 0.6 & 0.9 & 0.47 & 0.33 & 0.25\\\hline
Snappy & 1.6 & 0.4 & 0.7 & 0.21 & 0.17 & 0.13 \\\hline
zStd & 1.8 & 0.1 & 0.3 & 0.1 & 0.10 & 0.02 \\\hline
ndzip & 5.1 & 1.0 & 1.0 & 0.96 & 1.01 & 1.01\\\hline
\textbf{IBP} & \textbf{9.7} & \textbf{1.2} & \textbf{1.1} & \textbf{1.02} & \textbf{1.27} & \textbf{1.32} \\\hline
\end{tabular}
\end{table*}

\begin{table*}
\caption{Average space savings (\%) across algorithms/data.}
\label{tab:comp_ratio}
\centering
\begin{tabular}{|c||c|c||c||c|c|c|}  \hline
\multirow{2}{*}{\textbf{Algorithm}} & \multirow{2}{*}{\textbf{\shortstack{GNN\\Sparse}}} & \multirow{2}{*}{\textbf{\shortstack{GNN\\Dense}}}  & \multirow{2}{*}{\textbf{\shortstack{DLRM\\Weights}}} & \multirow{2}{*}{\textbf{\shortstack{LLM \\ KV (FP16)}}} & \multirow{2}{*}{\textbf{\shortstack{LLM \\ KV (BF16)}}} & \multirow{2}{*}{\textbf{\shortstack{LLM \\ Weights (BF16)}}}\\ 
 & & & & & & \\\hline
ANS & 76.7 & -78.1 & -132.6 & 6.2 & 18.84 & 12.40 \\\hline
Bitcomp & 80.5 & -8.8 & -14.9 & -0.15 & 0.59 & -0.97 \\\hline
Cascaded & 89.2 & -1.0 & -2.0 & -0.01 & 0.72 & -0.09 \\\hline
GDeflate & 86.9 & -22.7 & -40.8 & 33.25 & 34.24 & 19.20\\\hline
LZ4 & 89.7 & -0.3 & -1.0 & 33.33 & 27.63 & -0.30 \\\hline
Snappy & 88.6 & -1.0 & -1.0 & 23.24 & 11.64 & -0.73 \\\hline
zStd & \textbf{93.4} & 1.6 & -2.0 & \textbf{38.91} & \textbf{43.03} & 21.37\\\hline
ndzip & 62.9 & 6.3 & 4.8 & -0.21 & 4.95 & 4.31 \\\hline
\textbf{IBP} & 92.9 & \textbf{10.4} & \textbf{8.3} & 4.49 & 23.43 & \textbf{26.71}\\\hline
\end{tabular}
\end{table*}

\begin{table*}
\caption{Compression time (ms) across algorithms/data.}
\label{tab:comp_time}
\centering
\begin{tabular}{|c||c|c||c||c|c|c|}  \hline
\multirow{2}{*}{\textbf{Algorithm}} & \multirow{2}{*}{\textbf{\shortstack{GNN\\Sparse}}} & \multirow{2}{*}{\textbf{\shortstack{GNN\\Dense}}}  & \multirow{2}{*}{\textbf{\shortstack{DLRM\\Weights}}} & \multirow{2}{*}{\textbf{\shortstack{LLM \\ KV (FP16)}}} & \multirow{2}{*}{\textbf{\shortstack{LLM \\ KV (BF16)}}} & \multirow{2}{*}{\textbf{\shortstack{LLM \\ Weights (BF16)}}}\\ 
 & & & & & & \\\hline
ANS & 33.1 & 105.4 & 80.4 & 25.2 & 26.8 & 104.6 \\\hline
Bitcomp & 13.0 & 39.0 & 27.9 & 17.4 & 23.3 & 61.4 \\\hline
Cascaded & \textbf{12.8} & 33.6 & 26.7 & 35.7 & 46.9 & 120.8 \\\hline
Gdeflate & 68.5 & 482.7 & 316.8 & 394.2 & 420.9 & 1407.7\\\hline
LZ4 & 135.1 & 69.3 & 43.4 & 105.5 & 141.6 & 495.1\\\hline
Snappy & 133.3 & 81.8 & 55.3 & 165.2 & 181.0 & 440.9\\\hline
zStd & 53.5 & 292.1 & 266.3 & 121.5 & 150.1 & 515.7\\\hline
ndzip & 16.4 & \textbf{20.7} & \textbf{15.0} & \textbf{16.5} & \textbf{20.6} & \textbf{56.2}\\\hline
IBP & 151.5 & 53.2 & 34.2 & 42.8 & 40.9 & 94.9\\\hline
\end{tabular}
\end{table*}

To demonstrate this, we evaluate the overhead of nvCOMP v3.0.1~\cite{nvcomp}, a closed-source NVIDIA library that provides high-performance GPU implementations of popular compression algorithms, as well as ndzip-gpu~\cite{knorr:ndzip:2021}, an open-source GPU-accelerated compression algorithm shown to provide high decompression throughput for floating point data~\cite{Chen:FCBench:2024} frequently used in ML applications.
~\autoref{tab:comp_perf} shows the throughput of transferring and decompressing 100,000 compressed tensors, normalized to transferring uncompressed tensors on an A100 GPU, \autoref{tab:comp_ratio} shows the space savings, and \autoref{tab:comp_time} shows the amount of time taken to compress. 
We evaluate LLM compression across three scenarios: KV-cache in float16 (FP16) used by older models, KV-cache in bfloat16 (BF16) used by modern models, and weights in BF16.
See~\autoref{sec:eval} for details on experimental setup and datasets.
Historically, GPU FLOPs/bandwidth increase faster than PCIe bandwidth, retaining PCIe as the bottleneck~\cite{Memwall:Micro:2024}.

\begin{table*}[t]
\caption{Compression of bit packing compared to the fraction of 80p-invariant bits across datasets.}
\label{tab:invariant_bits}
\centering
\begin{tabular}{|c|c|c|c|c|c|c|c|c|c|}  \hline
\multirow{2}{*}{\textbf{Dataset}} & \multicolumn{3}{c|}{Sparse GNN} & \multicolumn{3}{c|}{Dense GNN} & \multirow{2}{*}{DLRM}  & \multirow{2}{*}{\shortstack{LLM\\(FP16)}}  & \multirow{2}{*}{\shortstack{LLM\\(BF16)}} \\ \cline{2-7}
 & Pubmed & Citeseer & Cora & Reddit & Product & MAG & & & \\\hline
\textbf{Bit pack \%} & 6.3\% & 6.3\% & 6.3\% & 0\% & 0\% & 0\% & 0\% & 0\% & 0\% \\\hline
\textbf{Group BP \%} & 7.5\% & 65.3\% & 76\% & 0\% & 0\% & 0\% & 0\% & 2\% & 0.1\%\\\hline
\textbf{Invariant \%} & 89.4\% & 99.0\% & 99.2\% & 15.0\% & 13.4\% & 12.9\% & 14.5\% & 8\% & 27\%\\\hline
\end{tabular}
\end{table*}

We find that many decompression algorithms perform worse than transferring uncompressed tensors, even though their space savings may be high. For example, zStd provides the best space savings for sparse GNN datasets, but its transfer speedup is one of the worst.
These algorithms frequently require multiple memory accesses to reconstruct each word in the decompressed tensor (e.g., dictionary lookups in LZ4), increasing PCIe usage. Simpler compression schemes, such as run-length encoding (RLE), delta coding, and simple bitpacking achieve a higher transfer speedup. \emph{Cascaded} employs a combination of them and attains the highest speedup among existing algorithms. However, these algorithms use metadata proportional to the size of data being compressed, leading dense datasets to have compressed tensors larger than the original tensors. The compression time also varies widely across these algorithms, with Bitcomp, Cascaded, and ndzip providing the shortest compression times.

\section{Towards Practical ML Data Compression}\label{sec:motivation}

To make lossless compression practical in ML workflows, it is essential to understand both the structure of input data and the architectural characteristics of GPUs. We analyze the patterns of invariance found in ML data, which form the basis for our compression (\S\ref{sec:invariance}). Then, we highlight the GPU-specific considerations that influence the design of an effective compression strategy (\S\ref{subsec:background_gpu}).

\subsection{Invariance in ML Data}\label{sec:invariance}

\begin{figure}
    \centering
    \includegraphics[width=0.4\linewidth]{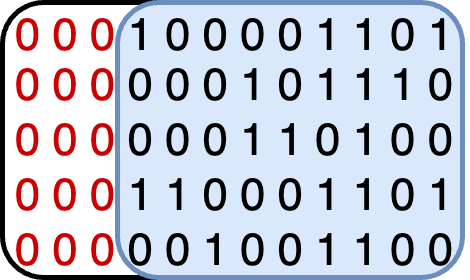}
    \hspace{3ex}
    \includegraphics[width=0.4\linewidth]{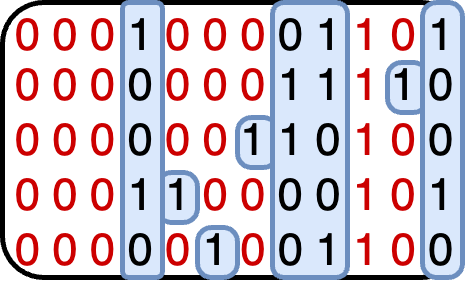}
    \Description[Conventional bitpacking and invirant bitpacking bits removed.]{Figures showing the first few bits removed for conventional bitpacking, and more bits removed for invariant bitpacking.}
    \caption{Conventional bit packing (left) and invariant bit packing (right). Red bits are removed to generate the compressed data shown in blue.}
    \label{fig:bitpacking}
\end{figure}

From our analysis in \S\ref{sec:compression_ml}, we find that bit packing performs best among existing algorithms. Bit packing is a family of compression algorithms commonly used in integer compression~\cite{bitpacking:SIGMOD:2013} to remove zero bits as shown in \autoref{fig:bitpacking} (left).
Algorithms such as Cascaded~\cite{nvcomp} and ndzip~\cite{knorr:ndzip:2021} make use of bit packing by performing a transformation on the input (e.g., delta coding).
These new data items are then packed into a smaller number of bits by removing zeroes, e.g., 12 to 9 bits in the figure.
Decompression is simple; as each compressed data item is known to be 9 bits long, offsets can be statically calculated, allowing GPU threads to parallelize decompression.
As only a single 9-bit data access is needed for each 12-bit item, repeated data accesses are avoided.

\smartparagraph{Limits of conventional bit packing}
Unfortunately, this method drastically limits the amount of compression we obtain.
To highlight this, we analyzed a few publicly available datasets.
\autoref{tab:invariant_bits} shows the percentage of bits that can be removed through bit packing for different datasets. 
As the range of values is large, the potential for space savings is very small.
One possibility to boost this is to segment the data into groups and perform bit packing within these groups, as done by ndzip and Cascaded. This works for some sparse GNN datasets (e.g., 6.3\% to 65.3\% for Citeseer with groups of 100 tensors). 
However, even for these sparse GNN datasets, compression is still limited.

\smartparagraph{ML data compresses well \emph{across} tensors}
We find that ML tensors tend to exhibit low entropy at the bit level, even when numerical values appear diverse.
Tensors are typically drawn from structured distributions rather than random bit patterns. In floating-point representations, values for a feature vector are often bounded due to regularization and training dynamics. Exponent and sign bits frequently have the same value across many different tensors. For example, for the Reddit GNN dataset~\cite{graphsage_reddit}, we found the second to fifth most significant bits (MSBs) had the same value more than 90\% of the time, and the first significant bit more than 75\% of the time. These correspond to the exponent and sign bits respectively. Conversely, the other bits shared a value roughly 50\% of the time, conveying a random distribution.
Sparse datasets have a bias towards zero-valued bits, which we saw in the Pubmed GNN dataset~\cite{pubmed_dataset}.
Generally, when values share a global structure, certain bit positions show low variance in values.

These bit positions are low entropy, where the probability of observing the same bit value (e.g., 0) is significantly higher than 50\%. When tensors are sampled from a stable distribution, such low-entropy bits remain consistent across tensors. As the number of sampled tensors increases, the empirical probability that such bits are invariant converges. We can exploit this property by identifying bit positions that are consistently fixed across a large fraction of tensors and storing them once globally. The remaining bits carry the entropy and are transmitted across PCIe normally.
Thus, compression is effective whenever tensors are drawn from structured distributions with shared low-entropy bit positions, as is common in GNN features, DLRM embedding tables, and LLM KV-cache/weight tensors.

We empirically demonstrate that ML data frequently has these low-entropy invariant patterns across tensors, (shown in \autoref{fig:bitpacking} (right)) by evaluating the invariant bits across different ML datasets/weights; \autoref{tab:invariant_bits} shows the fraction of bits that are invariant in 80\% of tensors or more (80p-invariant). 
Our notion of invariance is defined by value identity of the same tensor bit positions. We use a threshold value $T$ to state that a bit is invariant within an input set of tensors of cardinality $N$. 
80p-invariant means that at least $\frac{T}{N} = 80\%$ of tensors have identical bit values in the same position.
We see that all datasets have invariant bits.

Invariance persists across datatype sizes---MAG and LLM use FP16/BF16, the others use FP32.
Compared to FP16, BF16 LLM KV-caches contain significantly more invariant bits.
BF16 uses 8 exponent bits compared to FP16's 5, providing more opportunities for bits to be invariant.

To effectively reduce the amount of data ferried across PCIe, we wish to avoid transferring these invariant bits.
However, this presents multiple new challenges:
\circled{1} As each compressed tensor varies in size, fixed offsets are not possible, e.g., to decompress the fifth tensor bit, we would first have to parse and decompress the first four bits. This greatly hinders parallelization opportunities.
\circled{2} The bits are scattered, requiring multiple accesses to reconstruct the decompressed tensor. In \autoref{fig:bitpacking}, each tensor requires up to four separate bit accesses to decompress.

\subsection{GPU Architectural Considerations}\label{subsec:background_gpu}

Unlike CPUs, which can tolerate irregular memory access patterns and rely on complex control logic, GPUs achieve high throughput by executing thousands of lightweight threads in parallel. To fully leverage this parallelism, lossless decompression on GPUs must be designed to minimize memory latency, avoid thread divergence, and maximize coalesced memory accesses. Additionally, GPU memory is a constrained resource, and compression techniques must work within tight memory budgets while maintaining high decompression speed. In this section, we outline the key architectural features and performance considerations of modern GPUs that influence practical lossless compression for ML.

\smartparagraph{Hardware and software hierarchies}
The GPU's hardware is arranged in a hierarchy that allows for execution at a scale of hundreds of thousands of parallel threads. 
The main processor unit of a GPU is a \textit{Streaming Multiprocessor (SM)}~\cite{cuda_hardware}, with modern GPUs containing around a hundred SMs.
Each SM contains an L1 cache and \textit{shared memory} along with its execution units.
The shared memory is an L1 cache partition that is user-addressable.
The GPU memory is accessed by SMs through the shared L2 cache.
Multi-GPU setups are common, where multiple GPUs (typically up to 8) are connected by a high-bandwidth interconnect (e.g., NVLink~\cite{nvlink}).
This allows the different GPUs to access each other's memory to resemble a scaled-up GPU.
GPU programming environments, like CUDA~\cite{cuda_hardware}, expose a hierarchy of threads that mimic the hardware hierarchy. 
The smallest unit of execution is a thread. 
A group of 32 threads makes up a \textit{warp}, which executes in lockstep.
A \textit{thread block (TB)} is a group of warps that share the L1 cache and shared memory.
All warps in a TB are guaranteed to be within a single SM.

Leveraging this hierarchy is critical for high-performance GPU programs.
\textit{Warp primitives}~\cite{cuda_warp_prim} provide cycle-latency communication and synchronization among threads in the warp.
It is critical to dispatch GPU threads in a warp-parallel fashion to leverage these primitives.
Each warp can only execute a single common instruction at a time, so GPU programmers must ensure that threads within a warp execute the same code path to maximize throughput.

\smartparagraph{Optimizing CPU-to-GPU transfers}
Another critical aspect is the minimization of CPU interactions via the high-latency and bandwidth-constrained PCIe interconnect.
When CPU memory must be consulted, there are two methods to transfer data between the CPU and GPU: CPU-initiated and GPU-initiated.
In CPU-initiated transfers, the CPU programs the GPU's direct memory access (DMA) engine to transfer data to the GPU~\cite{cuda_transfer}.
While simpler to program, for small transfers (KBs) the overhead of programming the DMA engine dominates transfer time, lowering performance.

Alternatively, GPU-initiated transfers use GPU warps to directly access and move data from CPU memory to GPU via PCIe. This yields higher performance for small transfers by eliminating DMA and avoiding a PCIe round-trip to the CPU to start the transfer.
However, GPU-initiated PCIe transfers need to be carefully aligned to fully leverage the GPU warp parallelism. Aligned memory accesses can be \textit{coalesced} into a single transaction of up to 128B by a hardware coalescer~\cite{cuda_mem}. For unaligned accesses, the GPU hardware coalescer has to create multiple transactions for accesses across the alignment boundary, introducing overhead. Additionally, each PCIe transaction has header overhead, further reducing throughput. To maximize PCIe throughput, warps in a GPU should generate maximally aligned, 128B transactions when accessing CPU memory~\cite{emogi:VLDB:2020}.

\section{Invariant Bit Packing (IBP)}\label{sec:ibp}

To exploit invariance and solve these challenges, an ideal decompression technique for ML data must \circled{1} minimize the metadata required for compression, \circled{2} minimize memory accesses needed to decompress data, \circled{3} be easily parallelize to take advantage of the GPU, and \circled{4} support a diverse set of data types (e.g., float32, float16, bfloat16, sparse bitmasks).

We propose \textit{Invariant Bit Packing} (IBP) to achieve these goals. 
IBP consists of identifying invariant bits in data items, eliminating them, and shifting the remaining bits to reduce the space required for the data item. Decompression involves reading the input bits and reconstructing the original data by reinserting the removed bits. 
Invariant bit metadata is stored \textbf{once} in GPU memory and used for tensor decompression \textbf{across the entire dataset}.
Even for terabyte-sized datasets, the required metadata is extremely small, on the order of kilobytes.
This satisfies our first goal of minimizing the metadata for compression.
The key remaining challenge is to parallelize decompression while minimizing CPU memory accesses.
For GPUs, contiguous memory accesses by neighboring threads benefit from coalesced memory accesses, as threads co-located in the same processor share a common cache (\autoref{subsec:background_gpu}).
As IBP's packing is irregular, it is hard to predict where threads should access memory for decompression.

\smartparagraph{Warp-parallel iterative decompression}
We propose \emph{warp-parallel iterative decompression} for parallelization with minimal overhead while maintaining access locality.
First, we maintain a workspace in shared memory (i.e., a partition of the L1 cache).
The threads in the warp cooperate to read a contiguous segment from CPU memory to this workspace.
The data in the workspace can now be decompressed. 
However, each thread does not know from which bit to begin reading.
For this, they communicate using a warp primitive~\cite{cuda_warp_prim}, which provides cycle latency communication within a warp.
By performing this communication, the threads perform a prefix scan to ascertain their starting bit offset, then independently decompress the data items.
Once the entire workspace is decompressed, the next segment is read from CPU memory until the entire tensor is decompressed.
We make use of asynchronous memory accesses~\cite{cuda_async_transfer} to hide the overhead of decompression by overlapping it with CPU memory accesses.
The full process is described in \autoref{subsec:decomp}.

As the workspace is in the GPU shared memory, repeated accesses during decompression do not incur any PCIe overhead.
This ensures that the CPU memory is only scanned once for decompression, satisfying our second goal.
Dividing a tensor within a warp also increases parallelism, while minimizing communication overhead, satisfying our third goal.
Finally, IBP operates at a bit level, agnostic to underlying data types, and can be applied across a variety of types, e.g., float32, float16, int, or sparse vectors, fulfilling our last goal.

\begin{figure}[t]
    \centering
    \includegraphics[width=0.8\linewidth]{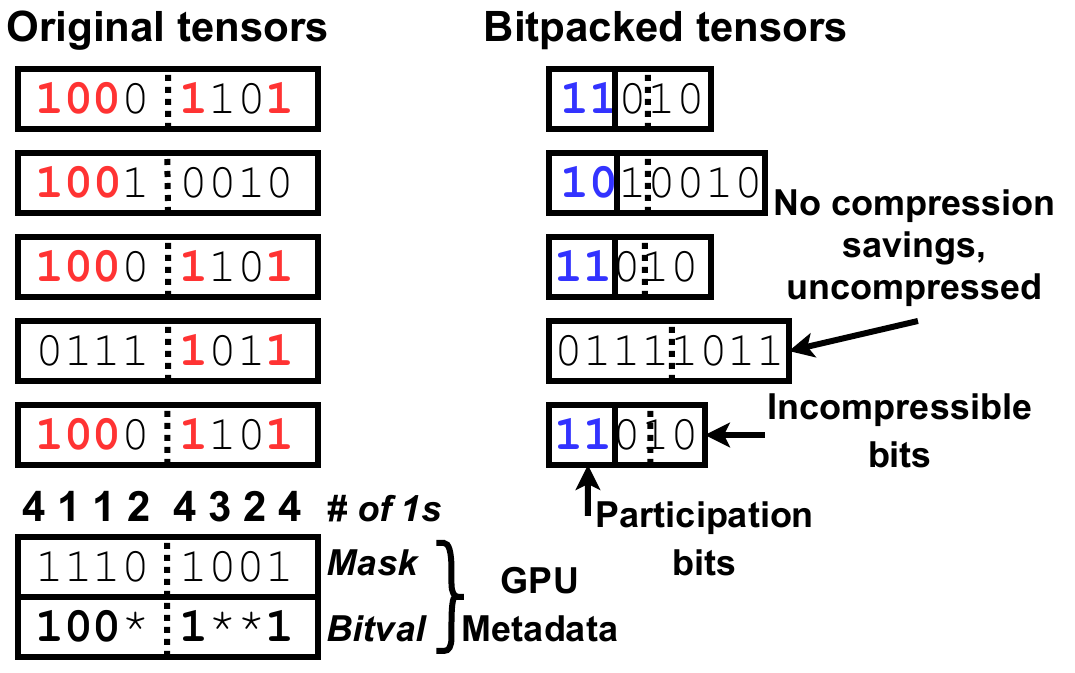}
    \Description[Diagram showing tensors being compressed.]{Matching bits in original tensors are highlighted. If all bits in the chunk match, these are removed showing the bitpacked tensors which are smaller in size.}
    \caption{Invariant bit packing example ($T=4$, $N=5$).}
    \label{fig:ibp_algo}
\end{figure}

\smartparagraph{Simple ML pipeline integration}
To simplify the integration of IBP in ML pipelines, we provide an easy-to-integrate open-source library.
We provide a PyTorch extension~\cite{pytorch_extension} for Python and CUDA support through a header-only library.
The Python functions are all called by the CPU, while the CUDA backend provides lower-level functions that can be called from either the CPU or the GPU.

We now look at how we concretely implement IBP,
referring to \autoref{fig:ibp_algo} which shows a simple example of IBP for $\frac{T}{N} = \frac{4}{5}$, with invariant bits shown in red. Typical input tensors are KBs in size; for exposition, we use 8-bit tensors.

\subsection{Preprocessing}

Before we can compress the input tensors, we must determine which bits are invariant. 
This step is performed on the GPU to exploit its parallelism.
Preprocessing is done once before model processing so that it does not interfere with the ML application.
For a $\sim180$GB dataset, preprocessing and compression only took 84 and 55 seconds respectively.

A counter is allocated for each bit across input tensors, e.g., 8 counters for 8-bit tensors or for each vertical column in \autoref{fig:ibp_algo}. For each set bit (i.e., value = 1) in each input tensor, we increment the corresponding counter for the bit---the count is shown underneath the original tensors.

From this, a \mask{} is constructed, containing a 1 for invariant bits, and a 0 otherwise.
\bitval{} contains the value of the invariant bits for each position.
If the value of a counter is more than $T$, that corresponding bit is set to 1 in the \mask{} and \bitval{}.
This implies that this bit is invariant with a value of 1.
If the value is less than $N - T$, where $N$ is the number of tensors in the dataset, the corresponding bit is set to 1 in the \mask{} and to 0 in the \bitval{}, implying the bit is invariant with a value 0. 
Otherwise, the bit in \mask{} is 0, implying the bit is not invariant.
For 0 bits in the \mask{}, the bits of \bitval{} do not matter, depicted with a *. 
We keep these bits as 0 in our implementation.
For the entire dataset, only one \mask{} and \bitval{} is constructed and used as metadata, which are stored in GPU memory due to their small size, avoiding large metadata overheads that scale with input.
For ML datasets, this gives good compression ratios (cf.~\autoref{subsec:eval:clustered}).

The value of $T$ has a heavy impact on the achieved compression ratio. 
If $T$ is too high, fewer bits get chosen as invariant, limiting the maximum achievable compression.
If $T$ is too low, excessive bits are chosen as invariant, leading to mismatches between tensors and \bitval{} values and causing many input tensors to remain uncompressed.
To maximize compression, we perform a sweep over $T$ values from $0.7N$ to $N$, picking the value providing the greatest observed compression.
In practice, $T = 0.8N$ strikes a good balance that maximizes compression for our datasets.
We explore the importance of $T$ further in~\autoref{subsec:eval_sens_chunk_threshold}.

\subsection{Compression}

Using \mask{} and \bitval{}, we compress the tensors.
Each tensor is first broken into fixed-size chunks (4 bits in the figure), allowing us to 
parallelize compression across chunks (and later, decompression), with each chunk handled by a different GPU thread.
A metadata bit is prepended to the compressed tensor for each chunk in the original tensor, called the \textit{participation bit} (blue in the figure).
The participation bit indicates whether that chunk is compressed.
A chunk can be compressed only if its masked bit values match the \bitval{}---formally, if
$Chunk_i \land Mask_i = Bitval_i$, for chunk $i$.
For chunks with matching values, the matching bits are removed in the compressed tensor (e.g., \texttt{100} for the first chunk in the figure), and the corresponding participation bit is set to denote that the chunk is compressed.
Otherwise, the participation bit is unset, e.g., as seen in the second chunk of the second tensor.

\begin{figure}[t]
    \centering
\begin{code}
def getCompressSize(tensor, tensorLen, Mask, Value):
  bitsSaved = 0
  i = THREAD_ID
  while i < tensorLen / CHUNK_SZ:
    chunk = tensor[i * CHUNK_SZ]
    maskChunk = Mask[i * CHUNK_SZ]
    bitvalChunk = Value[i * CHUNK_SZ]
    if (chunk BITAND maskChunk) == bitvalChunk:
      bitsSaved += popcount(maskChunk)
    i += WARP_SZ
  bitsSaved = warpScan(bitsSaved)
  # See if tensor is worth compressing
  if THREAD_ID == 0:
    participationBits = tensorLen / CHUNK_SZ
    bytesSaved = (bitsSaved - participationBits) / 8
    if bytesSaved <= 0:
      return tensorLen
    else:
      return tensorLen - bytesSaved

def compress(tensor, tensorLen, Mask, Value):
  j = THREAD_ID
  bitShift = 0
  while j < tensorLen / CHUNK_SZ:
    curBitshift = 32
    chunk = tensor[j * CHUNK_SZ]
    maskChunk = Mask[j * CHUNK_SZ]
    bitvalChunk = Value[j * CHUNK_SZ]
    if (chunk AND maskChunk) == bitvalChunk:
      curBitshift -= popcount(maskChunk)
    bitshift += warpScan(curBitshift)
    insertBits(outputTensor, chunk, bitshift)
    j += WARP_SZ
  return outputTensor

\end{code}
    \caption{Pseudocode for IBP tensor compression.}
    \label{fig:compression_code}
    \Description[Code for compression.]{Pseudocode illustrating functions and processes for compression.}
\end{figure}

Similar to preprocessing, compression is performed by the GPU to improve efficiency.
The tensors are retained within high-capacity CPU memory and accessed by the GPU using CUDA's zero-copy Unified Virtual Memory feature~\cite{cuda_uvm}, allowing the GPU direct access to CPU memory.
\autoref{fig:compression_code} shows the pseudo-code of the two main functions used for compression: pre-calculating a tensor's compressed size and the actual compression.
To calculate the compressed size of a tensor, we partition the set of all input tensors among the GPU warps, with each warp handling a subset of the input set. Each thread in the warp, in turn, is handling a separate chunk of its input tensor. The warps execute in parallel. For simplicity, we examine the operation of a single warp.

Each thread checks whether the bitwise AND operation between the chunk and \mask{} matches the \bitval{}, implying that the chunk can be compressed [lines 5-8].
If so, the number of set bits in the \mask{} indicates the number of bits saved in the chunk [line 9].
This process is repeated for the entire input tensor [lines 4, 10].
The warp then performs an in-place scan operation to determine the combined bits saved [line 11]. 
The leader thread of the warp [line 13] performs the final calculation of the compressed size.
The compressed tensor requires an extra bit per chunk for the participation bits, which are subtracted from the bits saved [lines 14 -- 15].
We keep all the tensors byte-aligned to avoid complex sub-byte memory accesses. Hence, we can divide the bits saved by 8 to get the bytes saved. 
If we find no bytes are saved, we keep the tensor in uncompressed form, without participation bits, returning the original size (e.g., the second-to-last tensor in the figure). In this way, the compressed dataset cannot exceed the size of the uncompressed dataset. In \autoref{sec:ibp_for_ml}, we shall see how we identify these uncompressed tensors. Otherwise, we return the new compressed size [line 18]. For the input tensors with compressed sizes smaller than the tensor size, we perform compression [line 21].

To compress the input chunk into the output buffer, each thread needs to know where to insert the compressed chunk.
For this, we calculate the \emph{bitshift} of the current chunk or the bit index at which the current chunk will be inserted.
This value depends on how many bits the previous chunks inserted into the output tensor, which are being compressed in parallel by other threads.
Waiting for them to complete serializes the operation, reducing compression performance.

To resolve this, we make use of \emph{warp intrinsics}~\cite{cuda_warp_prim}, which is a GPU feature that allows threads within a warp to communicate with each other with low latency (order of cycles).
Each thread first calculates the number of bits that its assigned chunk will occupy in the output tensor [lines 26 -- 30], and then all the threads communicate this value with each other using the warp intrinsic [line 31].
The threads then insert their chunk into the output array independently [line 32], maximizing parallelism.

\subsection{Decompression}\label{subsec:decomp}

\begin{figure}
    \centering
    \includegraphics[width=0.85\linewidth]{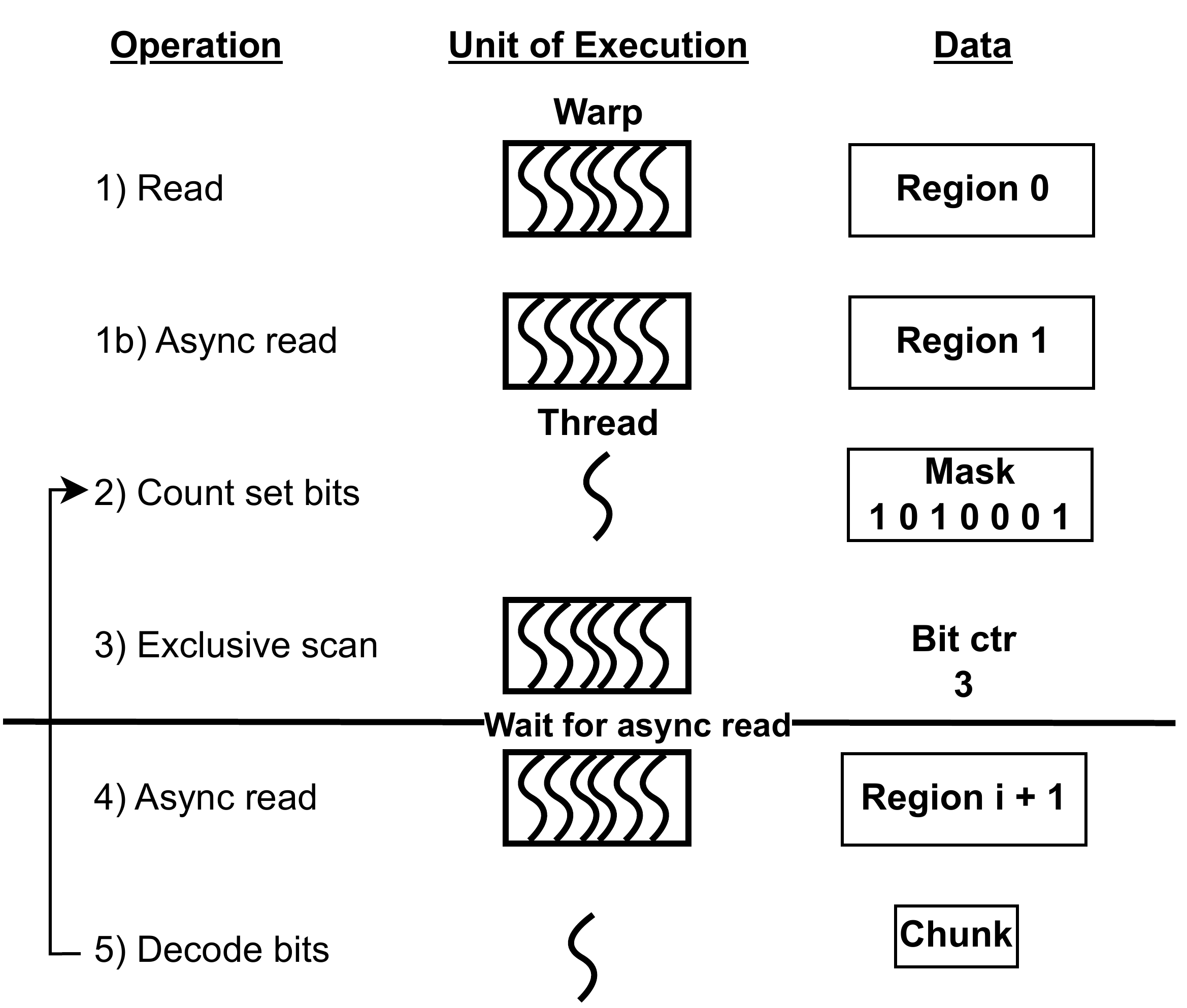}
    \caption{IBP tensor decompression steps.}
    \label{fig:decomp_steps}
    \Description[Step-by-step decompression diagram.]{First data is read by a warp, then another async read is done, then each thread counts the set bits in the mask. Then an exclusive scan counts bits for each thread. The previous async read is waited for before beginning decode.}
\end{figure}

Decompression is on the critical path of performance and must have minimal overhead.
We decompress input tensors in batches, with each tensor being decompressed by a separate warp, avoiding communication outside the warp and reducing synchronization overheads. 

\autoref{fig:decomp_steps} shows the major steps involved in decompression.
Each thread in the warp is responsible for decompressing a chunk within the tensor.
1) The warp collaboratively reads a region of data from the CPU memory to the workspace.
This collaborative read allows the operation to benefit from memory coalescing (\autoref{subsec:background_gpu}).
1b) Following this, an asynchronous read is initiated proactively.
2) Each thread then reads its respective chunk of the \mask{} metadata and counts the number of unset bits.
This indicates to the thread the number of bits that it will need to fetch from the shared workspace.
3) The warp then performs an exclusive scan on these bit counters using warp primitives to obtain individual offsets.
Each thread in the warp is now aware of how many bits the threads before it intend to insert, yielding the start offset for itself.
The warp then waits for the prior asynchronous read to complete.
4) If further CPU reads seem necessary, another asynchronous read is kicked off for the next region.
5) Each thread then reads the required bits from the workspace to decompress their assigned chunk.
This process is repeated until the entire tensor is decompressed.

At all points, communication is contained within the warp, reducing synchronization overheads.
CPU memory accesses are also asynchronous allowing the decompression logic to overlap with data movement.

\subsection{Efficient PCIe Data Transfer}\label{subsec:ibp_transfer}

\begin{figure}[t]
    \centering
    \includegraphics[width=\linewidth]{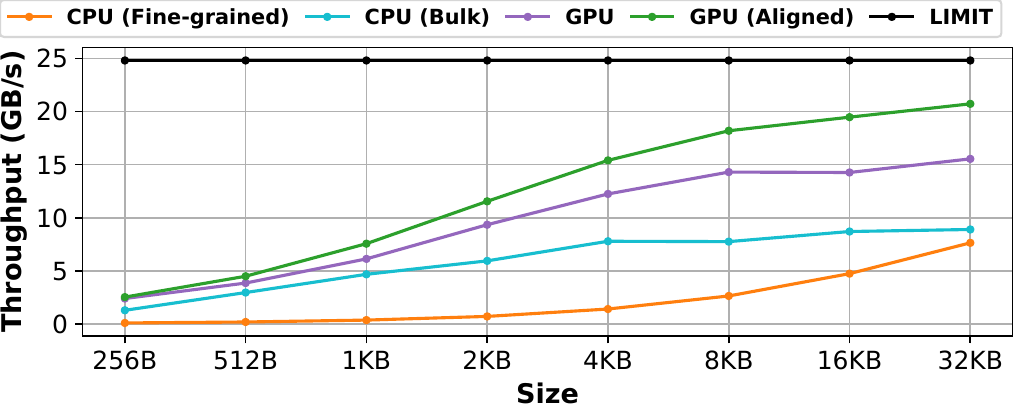}
    \caption{CPU-to-GPU copy throughput across methods.}
    \label{fig:copy_perf}
    \Description[Line graph of CPU-GPU copying performance.]{GPU (Aligned) does the best followed by GPU, then CPU (Bulk) and finally CPU (Fine-grained).}
\end{figure}

Efficient transfer of data across PCIe is crucial to reducing the overhead of decompression.
In \autoref{fig:copy_perf}, we evaluate four possible methods of copying randomly distributed, unaligned array elements of different sizes from CPU memory to a GPU buffer, along with the hardware limit. The first two are CPU-initiated. \emph{CPU (Fine-grained)} copies element-by-element to the GPU buffer. This performs the worst, as a DMA copy is performed for each array element, which has startup overheads. \emph{CPU (Bulk)} copies individual elements to a contiguous CPU buffer, then copies the CPU buffer at once to the GPU buffer. This performs better, as the DMA copy happens only once, amortizing the startup overhead. 
Systems like InfiniGen~\cite{infinigen:OSDI:2024} use this method to transfer tensors.
\emph{GPU} uses the CUDA zero-copy feature~\cite{cuda_uvm} to perform copying from within the GPU. This leverages the GPU's parallelism to transfer the array elements in parallel.

From this investigation, it is clear that initiating copies from the GPU performs better. However, as discussed in \autoref{subsec:background_gpu}, 128B size and aligned transfers provide maximum link utilization. Unfortunately, compressed tensors are misaligned and oddly sized, limiting performance. For this, we use two techniques: \circled{1} aligned and \circled{2} bounded data transfers.

\smartparagraph{Aligned transfers}
All our data transfers use aligned copies, where we find the 128B boundary for a CPU buffer being copied, and shift indices so that each warp starts at a 128B-aligned boundary.
We proceed with the accesses, copying the required data into a GPU buffer, with extra PCIe transactions only to copy unaligned portions.
\emph{GPU (Aligned)} shows this improves throughput by 25\% over simple GPU copying. 
Similar trends have been observed by other work~\cite{bamasplos, emogi:VLDB:2020}.

\smartparagraph{Bounded transfers} Our second technique, bounded data transfers, uses a lower bound and upper bound instead of a fixed length when copying. 
This guarantees aligned transfers when iteratively reading from a contiguous buffer.
The function uses a warp to copy 128B aligned segments from a source to destination, stopping when it has copied at least the lower bound up to the upper bound of data.

By ensuring the lower and upper bounds are separated by 128B, we copy data into an intermediate buffer, operate on it, and then copy more data while always copying 128B-aligned segments.
For example, Step 1 of \autoref{fig:decomp_steps} copies between 1--128 bytes based on the alignment of \verb+src+.
Without bounded transfers, if the source was misaligned and we repeatedly copied 128B into a buffer, every copy would be misaligned.

\section{Using IBP for ML acceleration}\label{sec:ibp_for_ml}
\begin{figure}
    \centering
    \begin{subfigure}{0.4\textwidth}
    \centering
    \includegraphics[width=\linewidth]{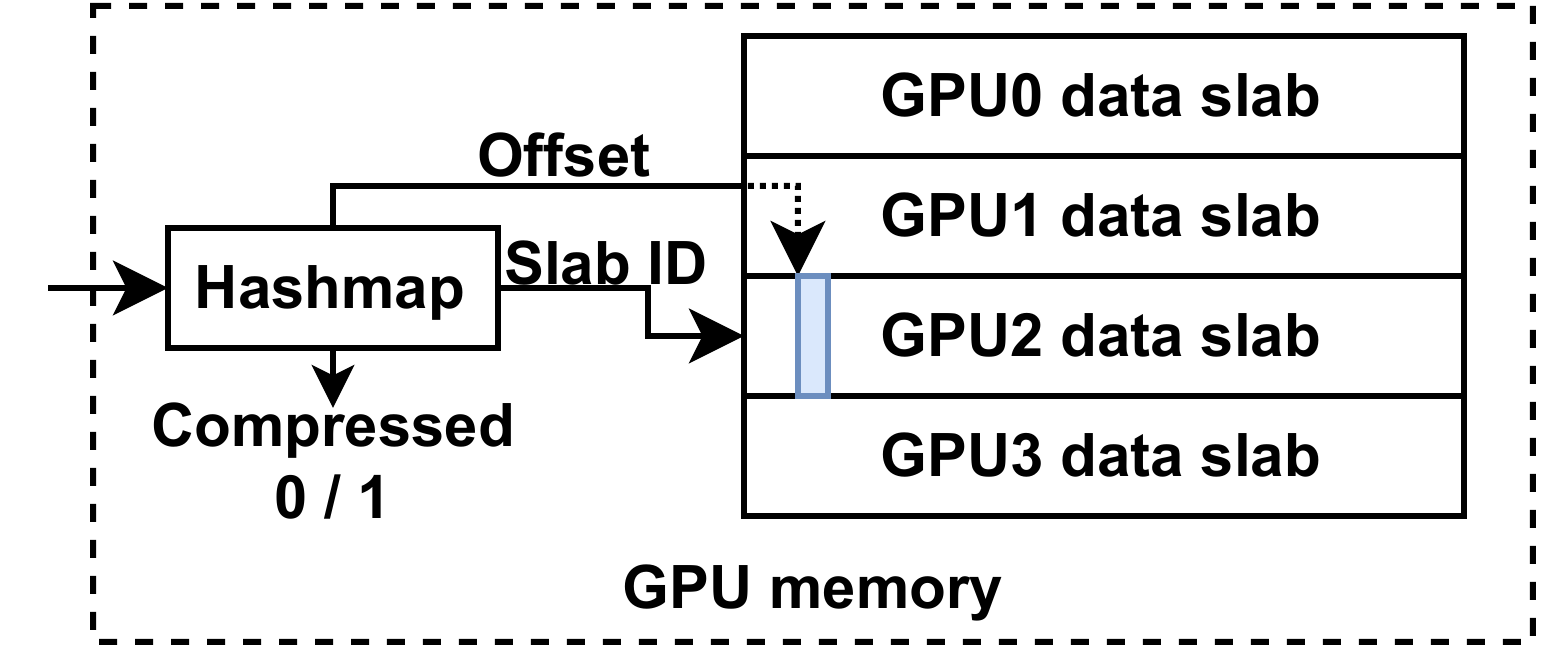}
    \caption{Compressed cache.}\label{subfig:ibp_ml_cache}
    \end{subfigure}
    \begin{subfigure}{0.4\textwidth}
    \centering
    \includegraphics[width=\linewidth]{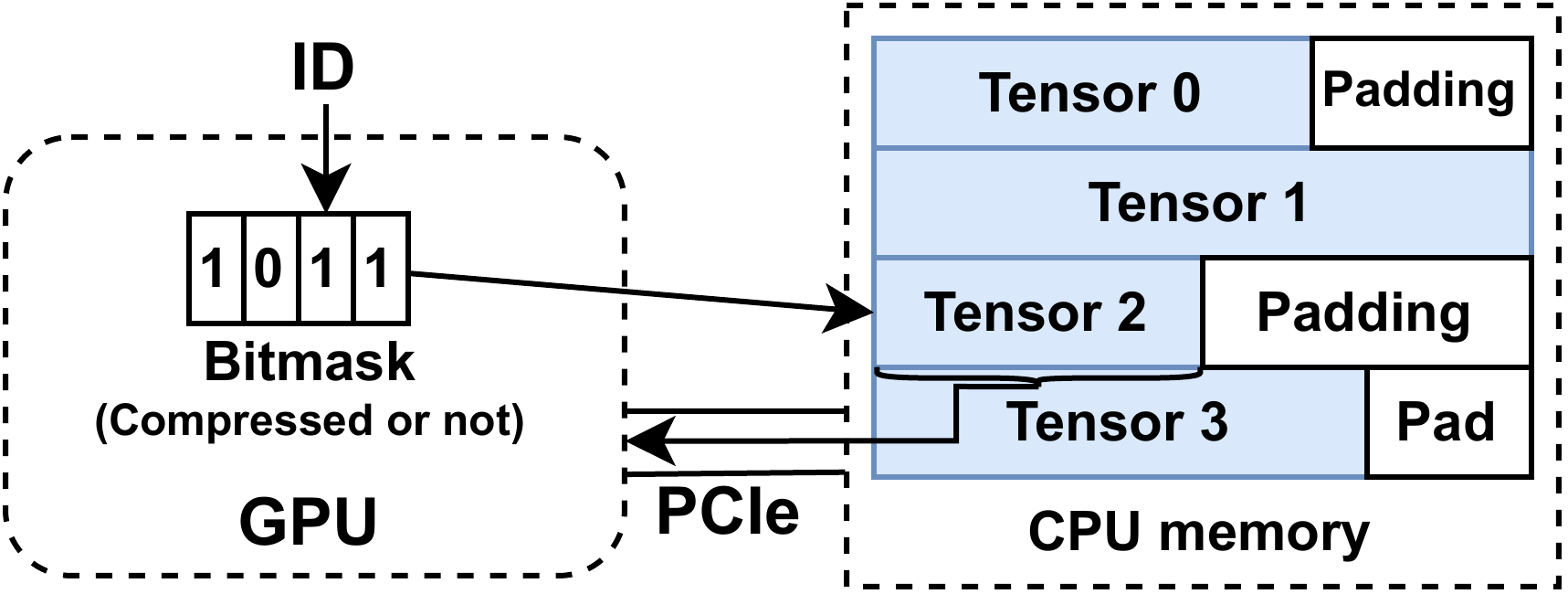}
    \caption{Compressed CPU memory.}\label{subfig:ibp_ml_mem}
    \end{subfigure}
    \caption{Integrating IBP into ML applications. }
    \label{fig:ibp_ml}
    \Description[Figures showing IBP integration.]{For caches, the hashmap outputs offset and slab ID providing details of where in GPU memory to check. For CPU memory, the tensors are padded to full size.}
\end{figure}

We now look at how we integrate IBP compression for ML acceleration, shown in \autoref{fig:ibp_ml}. 
We describe two systems; in the first (\autoref{subfig:ibp_ml_cache}), we compress a software cache in GPU memory to fit more tensors in the same space.
In the second (\autoref{subfig:ibp_ml_mem}), we compress tensors in CPU memory to reduce the amount of data ferried across PCIe.

\smartparagraph{Cache compression}\label{subsec:ibp_ml_cache}
To enhance a static GPU cache with compression, we must pack compressed tensors into the data store, adding new mappings into the hashmap.
To achieve this, we change the data store layout and hashmap output.

First, we allocate a contiguous memory region per interconnected GPU, which we call a data slab; e.g., an 8-GPU system would have 8 separate data slabs.
Together, these slabs make up the data store.
We compress the input tensors and lay them contiguously within the data slabs, keeping track of the offset within the slab and the GPU ID of the slab.
In this way, our compressed cache consumes the same amount of space as the uncompressed version.

During cache access, we require two pieces of information: where the tensor is located and whether it is in compressed or uncompressed form.
To avoid additional data structures or memory accesses, we integrate this information into the existing hashmap structure.
Instead of a 64-bit pointer to the relevant tensor, the hashmap now maps an ID to a 3-bit data slab ID, a 40-bit offset, and a bit indicating whether the tensor is in compressed form or not.
The 3-bit slab ID allows us to support up to 8 GPU slabs (the current multi-GPU limit), and the 40-bit offset allows us to index up to 1 TB of GPU memory, exceeding current GPU capacities.
We have 20 spare bits in case we need to extend these in the future. 

Using the data slab ID, we identify which GPU slab the required tensor is on and then use the offset to index into the appropriate location. The 1-bit compressed indicator then lets us know whether to decompress the data or simply read it. This retains all the existing features of the static cache, while seamlessly introducing compression.

\smartparagraph{PCIe transfer compression}\label{subsec:ibp_ml_mem}
While cache compression improves the hit rate of the cache, misses are still fetched from CPU memory.
To reduce PCIe transfer overhead, we also compress CPU tensors.
Two difficulties arise from this.
(1) For most datasets, input tensors are fixed in size and accessing a specific tensor involves shifting a pointer by a fixed offset. For example, to access the fifth tensor, one simply shifts 5 tensor indices.
The sizes of compressed tensors vary, and we cannot use a fixed shift to index a tensor.
Tracking offsets adds extra memory accesses when decompressing and increases memory usage.
(2) IBP retains certain tensors in uncompressed form, which must be separately tracked.

To solve the first problem, we perform in-place compression, where the entire dataset occupies the same amount of memory, but each individual tensor is compressed. 
For example, a 100B tensor may be compressed to 80B. The full 100B are allocated, but only the first 80B are useful.
When reading tensors from the CPU, only the required data is read, with individual tensors indexed as before. 
The CPU memory is many times larger than the GPU's, making this acceptable.

To solve the second problem, we maintain a bitmask of compressed and uncompressed tensors in GPU memory, which allows IBP to look up whether a tensor is compressed, and if so, decompress or otherwise simply transfer the entire tensor.
Each bit represents whether the CPU tensor is compressed or not. 
Before reading a tensor from the CPU, this bitmask is first checked. 
This has a memory overhead of 1 bit per tensor, or \textit{Bitmask size = Dataset size / (Tensor size * 8)}.
For a 512 GB dataset with 1 KB tensors, this bitmask occupies only 64 MB, fitting well within GPU memory. 

Using these techniques, we can reduce the amount of data transferred across PCIe by 8 - 93\% (\autoref{tab:comp_ratio}).

\section{Evaluation}\label{sec:eval}

In our evaluation, we wish to answer the following questions:
\begin{itemize}
    \item How well does IBP compress ML datasets versus the state-of-the-art? Does IBP achieve high performance and low overhead for decompression? (\autoref{sec:ibp_perf})
    \item What GNN training speedup can IBP provide? (\autoref{sec:gnn_speedup})
    \item What DLRM inference speedup can IBP provide? (\autoref{sec:dlrm_speedup})
    \item What is the LLM offload speedup with IBP? (\autoref{subsec:llm_offload})
    \item How do IBP's parameters impact performance? (\autoref{sec:ibp_sensitivity})
\end{itemize}

\smartparagraph{System configuration}
We evaluate IBP on a system with an A100 GPU with 300 GB of CPU memory.
Our system uses 16-lane PCIe Gen 4 as the CPU-GPU interconnect. 
This provides a theoretical bandwidth of 32GB/s per direction; we experimentally observed $\sim$25GB/s (\autoref{fig:copy_perf}).
We use CUDA 11.7, Python 3.8, and Pytorch 1.13.1. 

\smartparagraph{Baselines} 
For compression benchmarking, we use nvCOMP v3.0.1~\cite{nvcomp}, a closed-source NVIDIA library that provides GPU-accelerated compression algorithms, as well as ndzip-gpu~\cite{knorr:ndzip:2021}, an open-source GPU-accelerated compression algorithm for floating point data.
We integrate IBP into two GNN frameworks: DGLv2.1~\cite{dgl}, a popular GNN library, and Legion~\cite{Legion:ATC:2023}, a state-of-the-art (SOTA) GNN training framework.
For DLRM, we integrate IBP into Colossal-AI's CachedEmbeddingBag~\cite{colossolai:ICPP:2023, fang2022frequency}.
For LLM, we add IBP support to FlexGen~\cite{flexgen:ICML:2023}, where we offload LLM weights, and InfiniGen~\cite{infinigen:OSDI:2024}, where we offload the KV-cache.
We compare IBP against native systems without compression.

\subsection{IBP (De-)Compression Performance}\label{sec:ibp_perf}
We start by examining IBP's performance on real datasets, comparing it to existing GPU decompression methods. For compression, we are interested in the compression ratio. For decompression, we care about the throughput for compressed transfers across PCIe as a function of compression.

\smartparagraph{Dataset compression ratios}
\autoref{tab:comp_perf} shows that IBP provides the best transfer throughput. We now take a deeper look at the compression ratio that IBP provides and compare it to alternatives, shown in~\autoref{tab:comp_ratio}.
The sparse GNN datasets use sparse tensors.
For these, most algorithms are able to provide good compression, with zStd providing the best.
The dense GNN datasets, DLRM weights, and LLM KV-cache consist of dense floating-point numbers that are harder to compress.
zStd and ndzip can compress some of these. IBP is the only algorithm that compresses all of them, giving 10--12\% space savings for dense GNN datasets, 8\% for DLRM, and up to 27\% for LLM.

We experiment with different forms of LLM compression. We first compressed the KV-cache generated by an OPT-30B~\cite{metaopt:arxiv:2022} model, which yielded only 4 -- 5\% space compression.
This model uses the FP16 data type for KV-cache entries.

Modern models prefer the BF16 data type, which uses 8 bits for the exponent compared to the 5 exponent bits of FP16.
The expanded exponent allows the values to take a wider range, improving LLM accuracy.
As we note in \ref{sec:invariance}, exponent bits tend to be invariant, as values across tensors end up in similar ranges.

We test the Gemma-7B~\cite{gemma:2024} model which uses the BF16 type.
IBP delivers significant compression for both the LLM KV-cache and weights for this model at 23\% and 27\% respectively.
This benefit came with \emph{no changes} to IBP's algorithm, giving a strong demonstration of how it generalizes across data types and models.

Overall, IBP's compression across all datasets translates to practical throughput gains.

We also investigate the compression time in \autoref{tab:comp_time}. For IBP, this includes the time taken to preprocess the dataset to generate the \mask{} and \bitval{}. IBP requires a moderate amount of time to compress; we find that around 50--70\% of the time was spent on preprocessing the dataset, with the rest being shifting bits to compress the tensors. Generally, preprocessing is done off the critical path, reducing the significance of this overhead.

\smartparagraph{Decompression overhead}\label{sec:eval_ibp_perf}
We evaluate IBP's decompression overhead versus space savings across different tensor sizes via a microbenchmark. To achieve specific space savings, we create synthetic \mask{} and \bitval{} metadata, varying the set bits in the \mask{} to yield the desired compression ratio.
We compress 100,000 tensors of sizes representative of ML datasets (256B, 1KB, and 4KB) in CPU memory using the appropriate metadata and measure the total time taken to transfer and decompress the compressed tensors, averaged across 100 iterations.
To highlight decompression overheads, we normalize the throughput to that of a mock system that transfers compressed tensors across PCIe, but writes uncompressed tensors to GPU memory without decompression.

\begin{figure}
    \centering
    \includegraphics[width=0.9\linewidth]{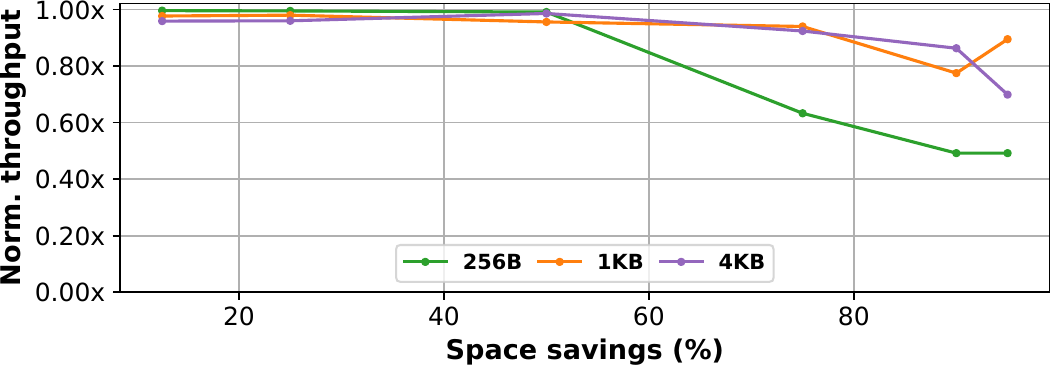}
    \caption{Decompression throughput versus space savings.}
    \label{fig:decompression_micro}
    \Description[Line graph comparing throughput and space saved.]{Up to 75\% space savings all remain near peak, whereas after they drop in throughput.}
\end{figure}

\autoref{fig:decompression_micro} shows that for space savings below 50\%, which covers the dense ML datasets, the decompression throughput is above $95\%$ of ideal. PCIe is the bottleneck in these cases, and the decompression overhead is hidden by the GPU's parallelism.
For compressed sizes smaller than this, i.e., 256B with 50\% or more compression, overhead is seen as excess data is brought in with a single 128B transfer.
As the space saved increases, throughput starts reducing as the overhead of decompression (e.g., GPU accesses, bitshifts) has an increasing impact on performance.
At 90\% space saved, which covers sparse datasets, throughput is at 78\% and 86\% of ideal for 1KB and 4KB, respectively, showing low overhead. Only small, 256B tensors suffer from an overhead of 50\%.

\smartparagraph{Dataset Sampling}\label{subsec:sampling}
It may be the case that the entire dataset cannot be preprocessed together, e.g., in distributed deployments or when data is updated at runtime or streamed in online scenarios.
For this, we can sample a fraction of the dataset for preprocessing.

In \autoref{tab:sens_sampling}, we generated a \mask{} and \bitval{} using a specified fraction of data instead of the entire dataset. We then evaluated the compression ratio obtained using this \mask{} and \bitval{} for the entire dataset.
This allows us to see how well a sampled \mask{} and \bitval{} generalizes to all data.
We see that samples generalize well to construct good \mask{} and \bitval{} for our datasets.
Even when 10\% of the dataset is sampled, the compression ratio obtained is comparable to the entire dataset in most cases.

\begin{table}
\caption{Compression ratios with different fractions of dataset sampled during preprocessing.}
\label{tab:sens_sampling}
\centering
\footnotesize
    \addtolength{\tabcolsep}{-2.5pt}
\begin{tabular}{|c|c|c|c|c|c|c|}  \hline
\textbf{Dataset} & \textbf{1\%} & \textbf{5\%} & \textbf{10\%} & \textbf{25\%} & \textbf{50\%} & \textbf{100\%}  \\\hline
Pubmed & 7.37x & 7.34x & 7.34x & 7.34x & 7.35x & 7.35x \\\hline
Citeseer & 25.09x & 25.09x & 25.09x & 25.09x & 25.09x & 25.09x \\\hline
Cora & 26.36x & 26.36x & 26.36x & 26.36x & 26.36x & 26.34x  \\\hline
Reddit & 1.13x & 1.13x & 1.13x & 1.14x & 1.14x & 1.14x  \\\hline
Products & 1.12x & 1.12x & 1.12x & 1.12x & 1.12x & 1.12x  \\\hline
MAG & 1.11x & 1.11x & 1.11x & 1.11x & 1.11x & 1.11x  \\\hline
DLRM & 1.11x & 1.12x & 1.13x & 1.13x & 1.13x & 1.14x  \\\hline
LLM KV (FP16) & 1.03x & 1.04x & 1.06x & 1.05x & 1.05x & 1.05x  \\\hline
\end{tabular}
\end{table}

\begin{table}[]
    \caption{Datasets used for GNN training. *SU are scaled up.}
    \label{tab:datasets}
    \footnotesize
    \centering
    \addtolength{\tabcolsep}{-4pt}
    \resizebox{\linewidth}{!}{
    \begin{tabular}{>{\centering\arraybackslash}m{1.3cm}|c|c|c|c|c|c}
    \textbf{Dataset} & PubmedSU & CiteseerSU & CoraSU & Reddit & Products & MAG \\ \toprule
    \textbf{Size per tensor} & 2KB & 14.5KB & 34KB & 2.4KB & 400B & 1.5KB \\\hline
    \textbf{Total tensor size} & 4.6GB & 34GB & 79.5GB & 0.52GB & 0.91GB & 174GB \\\hline\hline
    \textbf{No. of nodes} & 2.4M & 2.4M & 2.4M & 233K & 2.4M & 121M \\\hline
    \textbf{No. of edges} & 123M & 123M & 123M & 114M & 123M & 1.3B \\\hline
    \textbf{Dataset size} & 5.1GB & 34.5GB & 80GB & 0.95GB & 1.4GB & 180.2GB \\ \hline
    \textbf{Batch size} & 8,192 & 1,024 & 512 & 1,024 & 8,192 & 8,192 \\
    \end{tabular}
    }
\end{table}

\subsection{GNN Training Speedup}\label{sec:gnn_speedup}

\smartparagraph{Datasets}
We evaluate 6 public GNN datasets: Pubmed~\cite{pubmed_dataset}, Citeseer~\cite{citeseer_dataset}, Cora~\cite{cora_dataset}, Reddit~\cite{graphsage_reddit}, Products~\cite{ogb:Neurips:2020} and papers of MAG240M~\cite{ogb:Neurips:2020} (a superset of Papers100M~\cite{ogb:Neurips:2020}), spanning various tensor and dataset sizes (\autoref{tab:datasets}).

GNN training is commercially performed on datasets that span 100s of GBs to TBs in size~\cite{BGL:NSDI:2023}. 
Unfortunately, publicly available datasets are significantly smaller.
To compensate, we perform two modifications: 1) Pubmed, Citeseer, and Cora are very small datasets (order of MB); we substituted the original graph topology with the topology of Products to scale up the size of these datasets.
Input features from the original dataset were randomly assigned to the vertices of this larger graph.
We marked these datasets with the suffix \emph{SU} to indicate that they have been scaled up.
This retains a real-world graph topology, similar to scaled-up studies performed by prior work~\cite{Legion:ATC:2023, gnnlab:Eurosys:2022}.
As the features themselves are unmodified, scaling up does not affect the compression ratios obtained, which we empirically validated.
2) To imitate cache hit rates of large-scale training~\cite{BGL:NSDI:2023}, we restrict the GPU cache size to 1\% of the graph's total input tensor size.

\smartparagraph{Frameworks}
We extend DGL with our PyTorch extension to compress input tensors when transferring them from the CPU to the GPU.
Users call \verb+compress+ once during graph initialization to enable IBP.
IBP's one-time preprocessing and compression took between half a second to 3 minutes, depending on the dataset size.
After this, DGL's sampling process automatically decompresses input tensors after transferring them to the GPU.
Adding IBP to DGL required fewer than 35 additional lines of Python code, showing the simplicity of our Python API.

Legion is primarily written in CUDA and uses parallel processes for graph sampling and neural network training to overlap the two tasks.
It additionally makes use of a software-based static cache within the GPU memory to cache input tensors used for training and graph topology structures used for graph sampling.

We extend Legion by compressing the static cache to fit more input tensors within the same space.
We also compress the input tensors residing in CPU memory, decompressing them after transferring to reduce PCIe latency.

\smartparagraph{Setup}
We run the GraphSAGE model~\cite{graphsage_reddit} using two-hop random neighbor sampling with a fanout of 25 and 10.
We attempted to set a batch size of 8K used by prior works~\cite{Legion:ATC:2023, gnnlab:Eurosys:2022}, but ran into out-of-memory errors for certain datasets due to our feature tensors being larger than in those works; the evaluated batch sizes of datasets are shown in \autoref{tab:datasets}. 
We run 10 epochs of training for each dataset, and average the runtime of the last 9 epochs to avoid startup overheads seen in the first epoch.
IBP uses 4B chunk sizes as this provides good compression, while many of the dataset tensors are not divisible by an 8B chunk size.
We verified that IBP's lossless compression did not impact model accuracy.

\begin{figure}
    \centering
    \includegraphics[width=\linewidth]{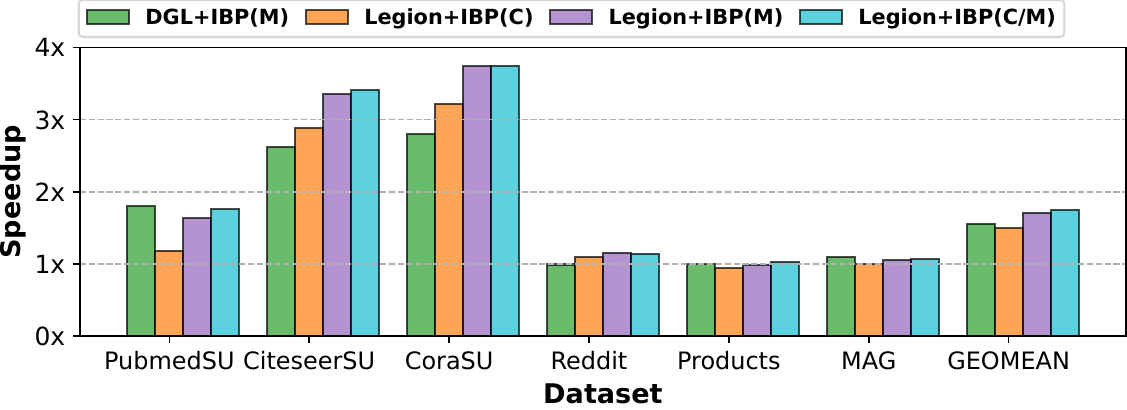}
    \caption{Average GNN training epoch speedup.}
    \label{fig:epoch_time}
    \Description[Bar graph showing performance]{IBP improves throughput when cache and memory is compressed across all datasets.}
\end{figure}

\begin{table}[]
    \caption{GNN tensor cache capacity with/out compression.}
    \label{tab:gnn_cache_cap}
    \footnotesize
    \centering
    \addtolength{\tabcolsep}{-4pt}
    \resizebox{\linewidth}{!}{
    \begin{tabular}{|c|c|c|c|c|c|c|}  \hline
    Cache & PubmedSU & CiteseerSU & CoraSU & Reddit & Products & MAG \\ \hline
Original & 24,401 & 24,401 & 24,401 & 2,301 & 24,401 & 1,217,501 \\ \hline
Compress & 166,901 & 589,182 & 631,609 & 2,596 & 26,987 & 1,354,043 \\ \hline
Increase & 6.84$\times$ & 24.15$\times$ & 25.88$\times$ & 1.13$\times$ & 1.11$\times$ & 1.11$\times$ \\ \hline
    \end{tabular}
    }
\end{table}

\smartparagraph{Performance}
\autoref{fig:epoch_time} shows the speedup of IBP over DGL and Legion, normalized to the respective framework. IBP(M) compresses the traffic across the PCIe, IBP(C) compresses the static cache, while IBP(C/M) does both. On average, IBP speeds up DGL by 56\% and Legion by 74\% on an A100.
The biggest gains IBP sees on top of Legion are due to PCIe traffic compression, improving throughput by up to $2.7\times$, and on average 70\%.
In general, having both cache compression and PCIe compression improves speedup over each individually.

We also look at how compression influenced the tensor cache used by Legion in \autoref{tab:gnn_cache_cap}. We see that the increase in capacity of the cache largely correlates with the ratio of compression of the datasets (\autoref{tab:sens_sampling}). More compressible datasets increase the capacity of the cache proportionately.

\begin{figure}
    \centering
    \includegraphics[width=\linewidth]{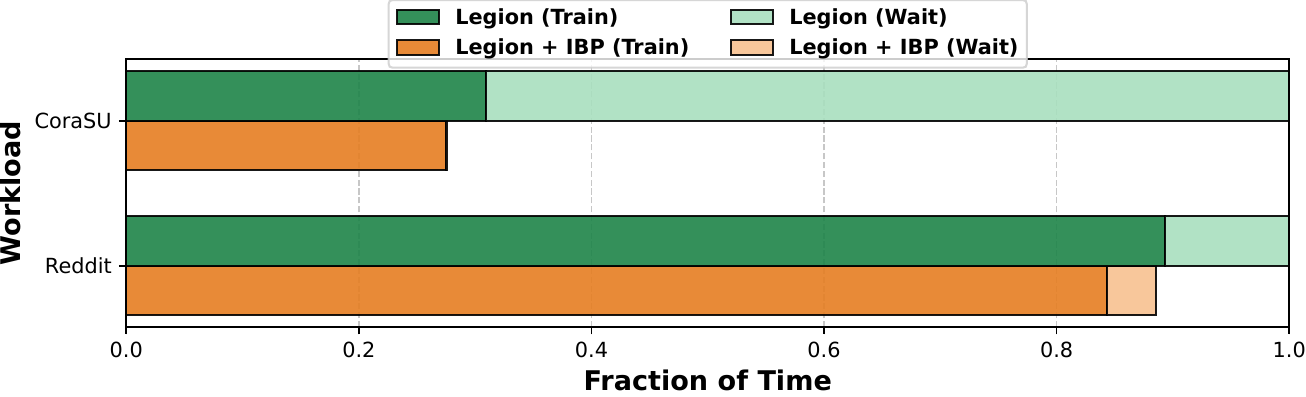}
    \caption{Breakdown of latency for GNN training epochs.}
    \label{fig:gnn_breakdown}
    \Description[Bar graph showing time spent training and waiting for data.]{For CoraSU, the baseline waits more than 50\% of the time. With IBP, this is completely removed. For Reddit, around 10\% of the time is spent waiting, which drops to a few percent.}
\end{figure}

\smartparagraph{Latency breakdown}
We examined the sources of latency in Legion, with and without IBP, for the CoraSU and Reddit datasets. We show these in \autoref{fig:gnn_breakdown}.
Legion uses two processes: one for computationally-heavy training, while the other samples the graph, prepares input, and fetches it into GPU memory.
These are done in parallel on two separate minibatches. The training process trains on a minibatch, while the graph sampler prepares the next minibatch.
We measured the time the training process spends training (Train) and waiting for the next batch (Wait). 
Wait shows the exposed latency of the sampler not hidden by Train.

We see that IBP reduces the Wait time greatly by reducing the amount of time spent transferring data.
For CoraSU, Wait time is almost eliminated entirely.
Interestingly, even the Train time slightly reduces. 
Legion performs data transfers using a subset of available SMs, with the remainder used for training.
IBP reduces data transfer volume, leading to a net reduction in transfer time.
Consequently, transfer SMs are free for ML training earlier, causing a cascading effect of speeding up Train time.
For CoraSU, IBP decreased training time by 12\% versus Legion, due to reduced SM contention for data transfer.

\begin{figure}
    \centering
    \includegraphics[width=0.9\linewidth]{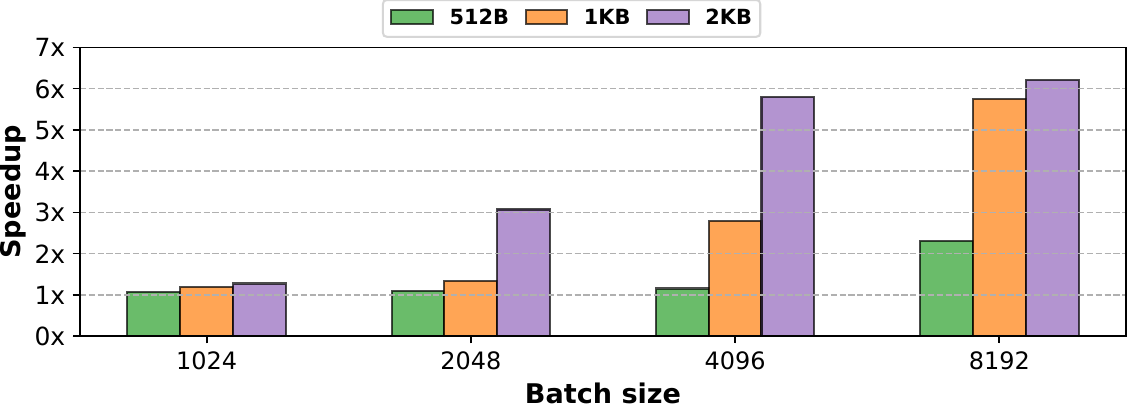}
    \caption{Normalized DLRM embedding lookup throughput with varying batch and entry sizes.}
    \label{fig:dlrm_thput}
    \Description[Bar graph showing DLRM speedup.]{IBP shows significant speedup for 2KB copies and at larger batch sizes.}
\end{figure}

\subsection{DLRM Inference Transfer Speedup}\label{sec:dlrm_speedup}

As discussed in \autoref{sec:pcie_in_ml}, large embedding table lookup and transfer is the bottleneck of DLRM inference~\cite{HugeCTR:RecSy:2022, Deeprecsys:ISCA:2020}.
Hence, we evaluate the benefit of IBP on embedding lookups.
Besides bookkeeping, adding IBP support required less than 15 lines of Python code.
The CachedEmbeddingBag maintains the embedding tables in CPU memory, caching frequently accessed embeddings in a buffer in GPU memory.
CachedEmbeddingBag supports both training and inference.
In training, the embedding table values can change, necessitating that evicted table entries from GPU memory have to be written back to the CPU.
For inference, these values are static. Hence, we slightly modify CachedEmbeddingBag to avoid unnecessary data movement from the GPU back to the CPU.
All other settings are kept to the default.

\smartparagraph{Setup}
We deployed embedding weights from a pre-trained model from NVIDIA~\cite{dlrm_weights} trained on the Criteo 1TB dataset~\cite{criteo_logs}, consisting of 26 embedding tables with an embedding entry size of 512B ($128\times$4B floats). 

The embedding tables vary greatly in size, with some being as small as 4 entries.
We focus on the large tables with at least 25,000 entries. Every lookup involves retrieving an entry from each of these tables.
We evaluate the embedding lookup time of CachedEmbeddingBag enhanced with IBP compared to the baseline, using different batch sizes.
To examine scalability, we rescale the embedding sizes from 512B to 2KB as industry models reach these larger sizes~\cite{swhwdlrm:ISCA:2022}.
Caching can help maintain frequently accessed data in GPU memory. However, DLRM tables have been shown to reach quite large sizes. As table size increases, cache hit rate decreases. Through our experiment we investigate how well IBP handles DLRM transfers.

\smartparagraph{Performance}
\autoref{fig:dlrm_thput} shows the throughput we obtained.
We found that the major source of improvement was due to the GPU-based data transfer that we employ (\autoref{subsec:ibp_transfer}).
In the baseline, the CPU performs the table lookups, which consume a significant amount of time. The looked-up entries are then transferred to the GPU.
As the batch and entry size increased, the overhead of CPU-based lookups increased.
Conversely, IBP provides a single Python function as a drop-in replacement for this operation, which performs the lookup and transfer in parallel on the GPU, aligning the transfers as needed.
This avoids the overheads of CPU lookups, using zero-copy to read directly from the GPU.

Compression reduced the overhead of transfers by 8\% to 20\% on average, reducing with increasing batch size.

\begin{figure}
    \centering
    \includegraphics[width=.9\linewidth]{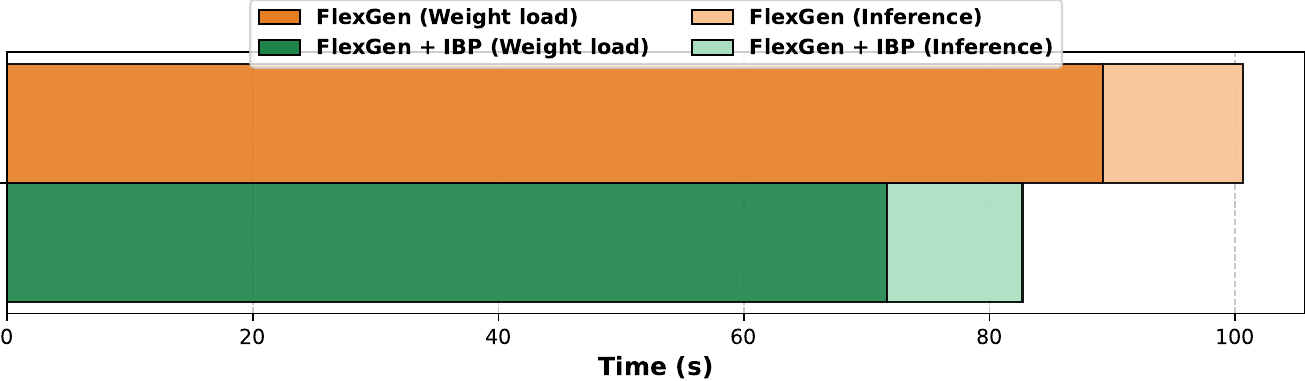}
    \caption{LLM inference latency with FlexGen weight offloading.}
    \label{fig:flexgen_comparison}
    
    \Description[Bar graph showing inference latency for FlexGen.]{IBP reduces latencies.}
\end{figure}

\begin{figure}
    \centering
    \includegraphics[width=.9\linewidth]{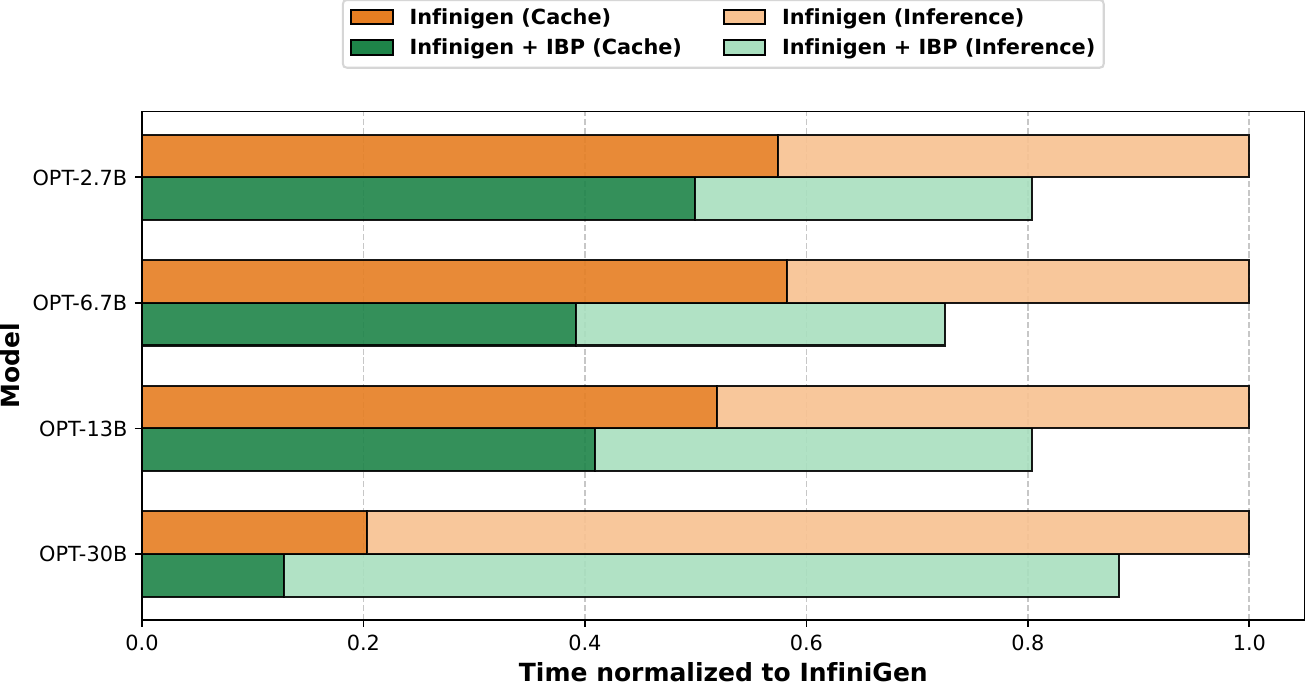}
    \caption{Normalized LLM inference latency with InfiniGen KV-cache offloading.}
    \label{fig:llm_comparison}
    
    \Description[Bar graph showing normalized inference latency for InfiniGen.]{IBP reduces latencies in all cases.}
\end{figure}

\subsection{LLM Inference Speedup} \label{subsec:llm_offload}
We test IBP's impact on LLM KV-cache and weight offloading.
We enhance two frameworks, FlexGen~\cite{flexgen:ICML:2023} and InfiniGen~\cite{infinigen:OSDI:2024}, with IBP to evaluate weight and KV-cache offloading respectively.
We use the same workload for both: a query of 1,920 input tokens with a batch size of 20, generating 128 output tokens.

Integrating IBP with FlexGen took $\sim20$ lines of Python code for the core logic.
The copy function, which transfers weights from the CPU to GPU, is replaced with IBP's decompress function, and compression is done while loading the weights to CPU memory.
We reserved only 64 SMs for decompression, allowing the rest to be used for concurrent inference computation.

We ran Gemma-7B~\cite{gemma:2024}, which uses BF16 weights.
These are stored in CPU memory, and brought layer-by-layer to the GPU.
\autoref{fig:flexgen_comparison} shows the inference latency, breaking down the time spent on weight loading and other inference latencies.
We see that IBP makes the end-to-end latency 22\% faster, driven by 25\% faster weight loading.
This improvement directly results from the compression that IBP gets with the BF16 LLM weights.

While the previous experiment demonstrated weight offloading, KV-cache transfers from the CPU also incur a heavy overhead.
To demonstrate this, we integrate IBP into InfiniGen~\cite{infinigen:OSDI:2024}, a recent LLM KV-cache management framework tailored for long-context inference that builds on FlexGen~\cite{flexgen:ICML:2023} minimizing KV-cache movement via indexing.
This required $\sim 40$ lines of code changes. It primarily involved replacing the function responsible for fetching the KV cache from CPU memory with our IBP-based implementation.
We reproduce an experiment from this work---we use Meta's OPT model~\cite{metaopt:arxiv:2022}, which uses the FP16 data type.We tried the newer Gemma model which uses BF16, but had issues getting it to work with InfiniGen due to conflicts with RoPE~\cite{rope:Neurocomputing:2024} operations.

For the KV-cache, we do not preprocess data for IBP.
Instead, we sample the prompt tokens' K/V to generate a \mask{} and \bitval{} during each prefill phase (cf.~\autoref{subsec:sampling}). This is used to compress the output token K/V entries during decode. The prefill time is slightly increased as a consequence and the associated overhead is included in our experiment. This increase is compensated by the reduction in transfer time.

\autoref{fig:llm_comparison} shows the KV-cache transfer and overall inference latencies with and without IBP. IBP reduces KV-cache movement latency by up to 37\% and reduces overall inference latency by up to 27\% versus InfiniGen. IBP preprocessing increases the input processing time by 10\%.
As input processing accounts for less than 20\% of total inference time, the gains in transfer speed outweigh this overhead.

The smaller models (2.7B, 6.7B, 13B) see larger inference latency reductions. OPT-30B does not fit in GPU memory and adds overhead for transferring uncompressed model weights, reducing the benefit of IBP’s KV-cache compression in the overall inference latency. The speedup is primarily due to IBP's GPU-initiated transfer (\autoref{subsec:ibp_transfer}). InfiniGen relies on the CPU to perform the KV-cache indexing and transfer, while IBP uses the GPU.
Compression improved transfers by 2\%.
InfiniGen transfers small 128B -- 256B K/V entries per head for OPT, and FP16 does not compress well, limiting the realizable transfer speedup per entry. 

Together, these experiments show IBP benefits LLM inference through two mechanisms: compression (dominant for BF16 weights) and efficient GPU-initiated transfer (dominant for sparse KV-cache).

\subsection{Performance Sensitivity Studies}\label{sec:ibp_sensitivity}
We now examine how performance varies with different configuration parameters and design choices.

\smartparagraph{Chunk Size (CS) and Invariant Threshold ($T$)}
\label{subsec:eval_sens_chunk_threshold}
Chunk size and invariant threshold as a fraction of considered tensors ($T/N$), have an impact on the compression and decompression performance that can be obtained.
A warp has 32 threads. To maximize parallelism, at least 32 chunks are desirable. However, each chunk requires a participation bit, affecting the compression ratio.
$T$ determines how often a bit needs to have the same value across $N$ tensors to be considered invariant, also affecting the ratio.

\begin{figure}
    \centering
    \includegraphics[width=0.85\linewidth]{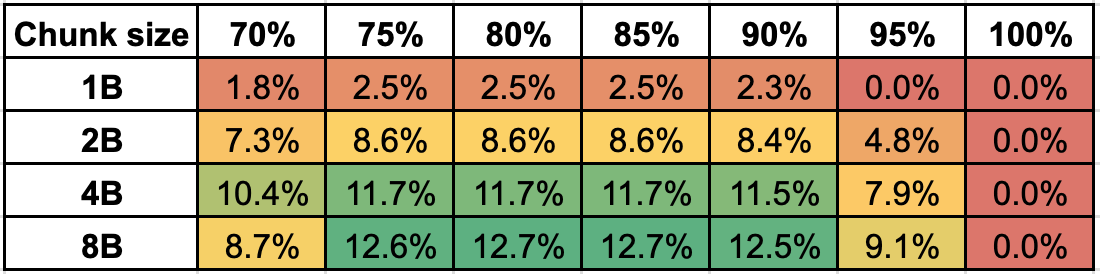}
\caption{Space saved with different chunk sizes and invariant threshold fractions $T/N$ for $N=$ 2.4M tensors in Products.}
\label{tab:sens_chunk_size}
    \Description[Colored table showing space saved.]{Peak savings are with 8B chunk size and 75 -- 90\% threshold fractions.}
\end{figure}

In \autoref{tab:sens_chunk_size}, we examine the effect of varying CS and $T/N$ on the Products dataset. For 1B CS (400 chunks), the compression ratio is low (< 3\%); this is because 1 participation bit is needed per byte, leading to 12.5\% overhead. As CS increases to 8B (50 chunks), this impact reduces to 1.6\%, increasing compression up to 12.7\%.
For high $T/N$ (e.g., 100\%), no bits are invariant, while a $T/N$ around 80--85\% strikes a good balance.
We evaluate varying $N$ in the next subsection.

\smartparagraph{Clustered Compression ($N$)} \label{subsec:eval:clustered}
So far, we have constructed a single \mask{} and \bitval{} for the entire dataset (setting $N$ to the cardinality of the dataset), exploiting globally invariant bits to provide compression.
This works well for the ML datasets we examined, but may be suboptimal for datasets with local trends among data items that are not globally prevalent.

For datasets with local invariance, a smaller $N$ can provide higher compression ratios. We call this \emph{clustered compression}, where the dataset is partitioned into $k$ clusters of size $N_c$ for cluster $c$, and a \mask{} and \bitval{} metadata is created for each cluster. How the dataset is partitioned into clusters has a major impact on compressibility. A common approach to finding similarity within a generic dataset is to use $k$-means clustering. This maximizes the probability that local invariance trends among tensors are exploited within each cluster, improving compression.
We modify $k$-means to use tensor bit-similarity for distances instead of absolute values.

\begin{figure}
    \centering
    \includegraphics[width=0.9\linewidth]{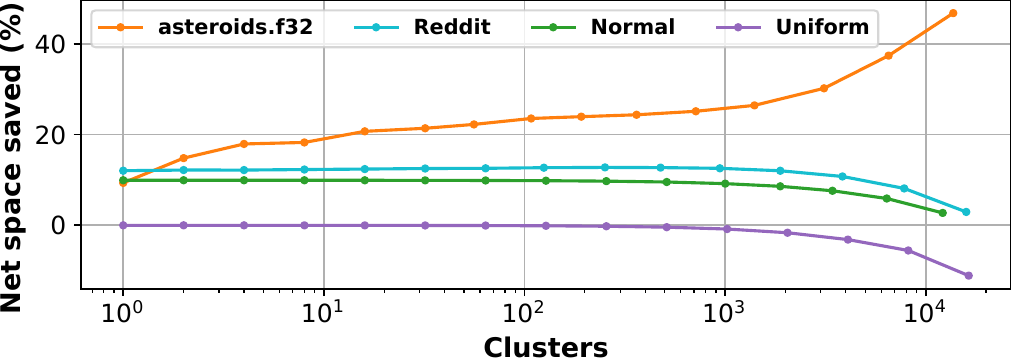}
    \caption{Clustered compression net space savings.}
    \label{fig:kmeans}
    
    \Description[Line graph showing space saved.]{asteroids dataset improves with clusters, while others diminish or stay relatively the same.}
\end{figure}

We demonstrate this approach in \autoref{fig:kmeans}, which plots the number of clusters $k$ (found via $k$-means clustering) versus the net space saved (i.e., space savings due to compression minus metadata overhead for additional \mask{} and \bitval{}) across a number of datasets, including datasets of 250,000 tensors containing normally and uniformly distributed random integers. We can see that our GNN datasets (Reddit) resemble a normal distribution, where clustering provides very minor benefit, i.e., the peak was 12.7\% savings with 477 clusters, compared to 12.0\% savings for one cluster. A uniform distribution is difficult to compress, as values have no trends to exploit.

To show that clustering can yield net space savings, we evaluate the asteroid.f32 dataset~\cite{ndzip_benchmarks, knorr:ndzip:2021}, containing the pressure component of a 3D simulation of an asteroid impact. The dataset contains a 3D tensor of $500\times500\times500$ floating point values, which we transformed into a set of 250,000 tensors of size 500.
The bit values of this dataset show local invariance trends --- as the number of clusters increases, these similar tensors get grouped together and the space savings increase.
With a single cluster, we obtain only $\sim 10\%$ net space savings. With 13,725 clusters, the dataset compresses by 58\%, yielding a net space saving of 47\% ($\sim11\%$ of the dataset size is consumed by \mask{} and \bitval{} metadata).

\begin{figure}
    \centering
    \includegraphics[width=.9\linewidth]{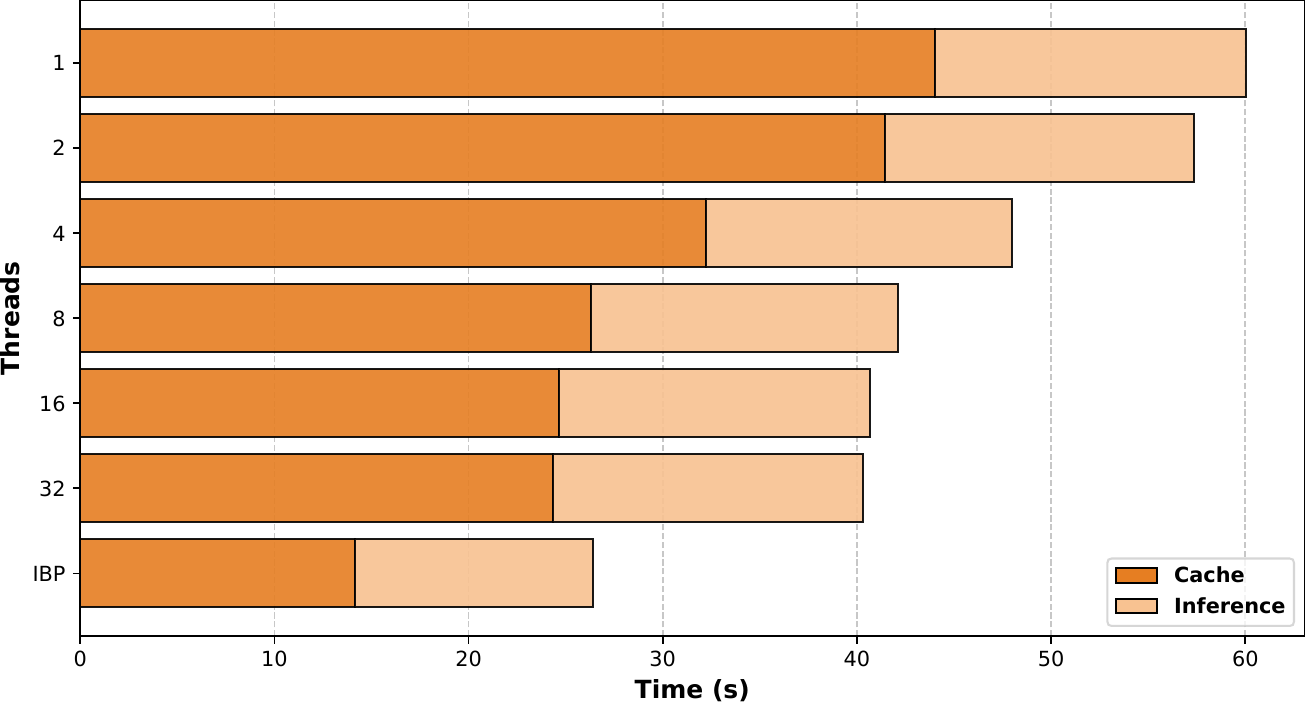}
    \caption{LLM inference latency with varying CPU threads.}
    \label{fig:llm_threads}
    
    \Description[Bar graph comparing inference latency with CPU threads.]{More threads decreases latency, but IBP's GPU-based copying beats all.}
\end{figure}

\smartparagraph{Analysis of CPU-based copying} \label{subsec:eval:cpu_copy}
Many existing frameworks rely on the CPU to copy data to/from the GPU, whereas IBP relies on the GPU.
We find that the performance of Pytorch's CPU-based copying varies with the amount of threads dedicated to copying.
We previously used 32 threads for our LLM and DLRM baselines.
In~\autoref{fig:llm_threads} we ran the LLM inference experiment varying the number of CPU threads dedicated towards Infinigen.

We see that as the threads increase, the KV cache management performance increases, due to more threads performing copying.
However, IBP's GPU-based copying still remains greatly superior.
In practical scenarios, dedicating large numbers of threads to copying is often infeasible.
The CPU performs many duties during ML inference, like fetching inputs~\cite{mlaas:NSDI:2022} or handling RPC calls~\cite{splitrpc:Sigmetrics:2023}.
Increasing the number of CPUs for tasks like copying often moves the bottleneck from PCIe transfers to CPU contention.
Conversely, as the GPU is waiting on transfers, making use of its waiting SMs does not increase contention, while alleviating the PCIe transfer bottleneck.

\section{Related Work}
\smartparagraph{GPU-accelerated compression} 
Several works propose GPU-based compression~\cite{int_comp:SIGMOD:2022, knorr:ndzip:2021, bzip2_gpu:ICPP:2024, nvcomp, mpc:ICCC:2015, bzip_gpu:inpar:2012, rANS:ICPP:2023} focused on exploiting the GPU's parallelism and high throughput to provide high compression and decompression throughput, including bit packing~\cite{int_comp:SIGMOD:2022, knorr:ndzip:2021}.
Other work~\cite{FastLanes:Damon:2024, giddy:damon:2017} also examined GPU-accelerated compression in the context of integer-heavy database workloads.
NVIDIA A100 and newer GPUs also support generic lossless hardware GPU memory compression~\cite{compressible_mem}. 
On the opposite end, many projects~\cite{mlcomp1:PMLR:2017, mlcomp2:cvpr:2020, mlcomp3:cvpr:2019, mlcomp4:neurips:2023} explore using ML to improve lossy compression performance.
Unlike these works, IBP applies bitpacking to ML datasets and CPU-to-GPU transfers, trading off preprocessing overheads for fast, lossless GPU tensor decompression.

\smartparagraph{Exploiting sparsity in ML} 
Many works try to reduce sparsity in ML applications. Quantization~\cite{qlora:Neurips:2023, bifeat:arxiv:2023, fang2022frequency, dlrm_quant:SC:2024, atom:mlsys:2024,dlrm_quant:neurips:2022} is a popular lossy method which trades off precision for reduced space.
Compression in ML has also been investigated in the context of gradient compression~\cite{grace,cnat,dgc,thc} and collective communication~\cite{wang2024zero,omnireduce}. Typically, compression is lossy and can affect the final evaluation results.
Various works~\cite{sparGNN:ASPDAC:2024, sgcn:HPCA:2023,growgnn:HPCA:2023} have also proposed hardware accelerators for GNN training, making use of hardware compression to reduce data accesses. 
CacheGen~\cite{CacheGen:SIGCOMM:2024} investigates the network for LLM KV transfers, proposing dynamic compression to meet latency goals under fluctuating network bandwidths.

Other sets of work~\cite{sparse_mat:TPDS:2013, cla:vldb:2016, sgd:sigmod:2019, aware:mod:2023} look at compressed matrix algebra to improve ML workload performance. FlexPoint~\cite{flexpoint:neurips:2017} proposes a reduced-size floating point data format, similar to TensorFloat32~\cite{tensorfloat32} and BFloat16~\cite{brainfloat16} seen in commodity GPUs today. IBP instead focuses on tensor compression for commodity GPUs with limited code change to speed up CPU-GPU data transfers.

DFloat11~\cite{dfloat11:Neurips:2026} and ZipServ~\cite{zipserv:ASPLOS:2026} show the benefits of lossless compression for LLMs.
They use statistical trends in LLM weight values to provide solutions tailored to the bfloat16 data type, targeting LLM weight compression.
Unlike these, IBP is a general algorithm meant for any data type and ML tensor, e.g., extending to KV-cache entries as well.

\section{Conclusion}

We propose lossless compression for ML to eliminate invariant bits across tensors, without compromising data fidelity. Our proposal, IBP, uses GPU-accelerated decompression to minimize overhead. Easy integration is demonstrated through IBP-enabled implementations within GNN, DLRM, and LLM frameworks, observing significant performance improvements. This demonstrates IBP’s potential to alleviate GPU memory capacity and PCIe bottlenecks, enabling scalable and efficient ML workflows.

\section*{Acknowledgments}

We thank our shepherd Yuke Wang and the anonymous reviewers for their valuable feedback on the paper and artifact. We thank Amandio Faustino for his invaluable assistance with system setup, debugging, and facilitating access to the computing environment used in this work. This work is supported by NSF grants 2148209 and 2212193, as well as the University of Washington Center for the Future of Cloud Infrastructure (FOCI).
Portions of this work were revised using GenAI tools, in compliance with ACM policy.

\bibliographystyle{plain}
\bibliography{references}

%%%%%%%%%%%%%%%%%%%%%%%%%%%%%%%%%%%%%%%%%%%%%%%%%%%%
% Artifact Appendix Template for EuroSys'26 AE
%
% this document has a maximum length of 2 pages.
%%%%%%%%%%%%%%%%%%%%%%%%%%%%%%%%%%%%%%%%%%%%%%%%%%%%

\appendix
\section{Artifact Appendix} 

%%%%%%%%%%%%%%%%%%%%%%%%%%%%%%%%%%%%%%%%%%%%%%%%%%%%%%%%%%%%%%%%%%%%%
\subsection{Abstract}
We provide the source code and setup necessary for Invariant Bit Packing library alongside applications and microbenchmarks necessary to replicate the results from this paper.

The library contains the CUDA backend and Python module for IBP. The applications are GNN (DGL~\cite{dgl} and Legion~\cite{Legion:ATC:2023}), DLRM~\cite{dlrm_weights}, and LLM~\cite{infinigen:OSDI:2024}. All necessary datasets are directly downloaded by our provided scripts.

%%%%%%%%%%%%%%%%%%%%%%%%%%%%%%%%%%%%%%%%%%%%%%%%%%%%%%%%%%%%%%%%%%%%%
\subsection{Description \& Requirements}
Our requirements are primarily hardware, specifically an NVIDIA A100 GPU. The software requirements are handled through Docker.

\subsubsection{How to access}
\paragraph{GitHub:} \url{https://github.com/AKKamath/InvariantBitPacking}

\paragraph{Zenodo:} \url{https://zenodo.org/records/18869047}

\subsubsection{Hardware dependencies}
An NVIDIA A100 GPU and minimum 300 GB CPU memory, 1 TB disk space.

\subsubsection{Software dependencies}
The software dependencies will get resolved in our Docker installation.

We tested on Ubuntu 22.04 OS with CUDA 11.7, Python 3.8, and Pytorch 1.13.1.

\subsubsection{Benchmarks} 
All required benchmarks are downloaded directly by the scripts in the artifact.
These are: Pubmed~\cite{pubmed_dataset}, Citeseer~\cite{citeseer_dataset}, Cora~\cite{cora_dataset}, Reddit~\cite{graphsage_reddit}, OGB-Products~\cite{ogb:Neurips:2020}, and MAG~\cite{ogb:Neurips:2020} GNN datasets; DLRM precomputed weights~\cite{dlrm_weights}; and OPT-2.7B, 6.7B, 13B, and 30B LLM models~\cite{metaopt:arxiv:2022}.

%%%%%%%%%%%%%%%%%%%%%%%%%%%%%%%%%%%%%%%%%%%%%%%%%%%%%%%%%%%%%%%%%%%%%
\subsection{Setup}
The repository contains a README.md file with instructions on setup. 
That is the definitive instruction on setting up, and may be more up-to-date than instructions given here.
\\

\noindent Download repository and datasets:
\begin{footnotesize}
\begin{lstlisting}[frame=single,rulecolor=\color{black}]
git clone https://github.com/AKKamath/InvariantBitPacking.git
cd InvariantBitPacking
git submodule update --init --recursive

make download_dlrm # 16GB download
make download_llm # 8GB download
\end{lstlisting}
\end{footnotesize}

We provide two methods of installing and testing, Docker and manual installation. We highly recommend using the Docker approach, as the manual installation can run into issues with library version mismatches causing errors.

\subsubsection{Docker} We provide a docker image with all dependencies pre-installed. You need to have \href{https://docs.docker.com/engine/install/}{Docker} and \href{https://docs.nvidia.com/datacenter/cloud-native/container-toolkit/latest/install-guide.html}{NVIDIA Container Toolkit} installed on your system. You can build the docker image and launch as follows: 
\begin{footnotesize}
\begin{lstlisting}[frame=single,rulecolor=\color{black}]
# Build the local docker image
docker build -t ibp-image -f Dockerfile .
# Run the docker image
docker run --gpus all -it  -p 8181:8181 \
    --rm --ipc=host --cap-add=SYS_ADMIN ibp-image
\end{lstlisting}
\end{footnotesize}

\subsubsection{Manual installation}
For manual installation, we download the IBP repository to the home directory and install it. We use Anaconda for fetching the appropriate versions of CUDA, Python, and PyTorch. This can take up to 2 hours.
\begin{footnotesize}
\begin{lstlisting}[frame=single,rulecolor=\color{black}]
# If conda not installed:
make install_miniconda
# Setup:
make create_env
conda activate ibp
# If system does not have CUDA 11.7 installed, the below installs it in conda.
make install_cuda
conda env config vars set CUDA_HOME="${CONDA_PREFIX}"
conda env config vars set CUDA_TOOLKIT_ROOT_DIR="${CONDA_PREFIX}"
conda env config vars set CUDACXX="${CONDA_PREFIX}/bin/nvcc"
conda env config vars set PATH=$CONDA_PREFIX/bin:$PATH
conda env config vars set LD_LIBRARY_PATH=$CONDA_PREFIX/lib:$LD_LIBRARY_PATH
# If CUDA 11.7 is already installed, continue from here.
make install_deps
conda deactivate
conda activate ibp
# Install NVComp, NDZip, Legion, DGL, ColossalAI, IBP
make install
# Download GNN dataset (dependent on DGL)
make download_gnn
\end{lstlisting}
\end{footnotesize}

%%%%%%%%%%%%%%%%%%%%%%%%%%%%%%%%%%%%%%%%%%%%%%%%%%%%%%%%%%%%%%%%%%%%%
\subsection{Evaluation workflow}

\subsubsection{Major Claims}
The claims made in the paper follows:\\
\begin{itemize}
    \item (C1): IBP achieves the best average speedup of compressed CPU-to-GPU transfers across all datasets as shown in \autoref{tab:comp_perf}.
    \item (C2): IBP achieves speedups for DGL and Legion for GNN training as shown in \autoref{fig:epoch_time}.
    \item (C3): IBP improves throughput for DLRM embedding lookups as shown in \autoref{fig:dlrm_thput}.
    \item (C4): IBP reduces inference latency for InfiniGen as shown in \autoref{fig:llm_comparison}.
\end{itemize}

\subsubsection{Experiments}
The instructions below explain how to run the experiments to evaluate the major claims listed prior. Due to system differences and runtime variation, the exact performance numbers may not match those seen in the paper, but the relative trends should hold true.

\textbf{Experiment (E1):} \textit{Comparison of different compression algorithm performances.} \\
This experiment compares the relative speed and compression ratios of the different algorithms on the different datasets.
To run this experiment type `\lstinline|make nvcomp_comparison|'. 
The results will be found in results/nvcomp\_comparison.log; you can run `\lstinline|cat results/nvcomp_comparison.log|' to output to terminal. 
\textbf{Note that the output provides the details for every dataset individually.} For the sake of space, \autoref{tab:comp_perf} and \autoref{tab:comp_ratio} considers the average of Pubmed, Citeseer, and Cora as `Sparse GNN', and the average of Reddit, Product and MAG as `Dense GNN'.
This should verify claim C1 by generating results similar to \autoref{tab:comp_perf} and \autoref{tab:comp_ratio}.

\textbf{Experiment (E2):} \textit{GNN training performance.} \\
This experiment compares the runtime of DGL and Legion with and without IBP enabled.
To run this experiment type `\lstinline|make gnn|'. 
The results will be found in results/gnn
\_perf.log. 
You can run `\lstinline|cat results/gnn_perf.log|' to output to terminal, then copy-paste into your favorite graph plotting software (e.g., Excel, Google Sheets) to plot into a bar chart. 
This should verify claim C2 with results similar to \autoref{fig:epoch_time}.

\textbf{Experiment (E3):} \textit{DLRM lookup throughput.} \\
This experiment compares the lookup throughput of DLRM embeddings with varying batch and entry sizes.
To run this experiment type `\lstinline|make dlrm|'. 
The results will be found in results/dlrm\_perf.log.
You can run `\lstinline|cat results/dlrm_perf.log|' to output to terminal, then copy-paste into your favorite graph plotting software (e.g., Excel, Google Sheets) to plot into a bar chart. 
This should verify claim C3 with results similar to \autoref{fig:dlrm_thput}.

\textbf{Experiment (E4):} \textit{LLM inference latency.} \\
This experiment compares the inference latency of InfiniGen with and without IBP enabled.
To run this experiment type `\lstinline|make llm|'. 
The results will be found in results/llm\_latency.log. 
You can run `\lstinline|cat results/llm_latency.log|' to output to terminal, then copy-paste into your favorite graph plotting software (e.g., Excel, Google Sheets) to plot into a bar chart. 
This should verify claim C4 with results similar to \autoref{fig:llm_comparison}.

%%%%%%%%%%%%%%%%%%%%%%%%%%%%%%%%%%%%%%%%%%%%%%%%%%%%%%%%%%%%%%%%%%%%%
\subsection{Notes on Reusability}
\label{sec:reuse}
The paper proposes a methodology on transferring compressed data during execution time of ML models to reduce the transfer overhead. 
We provide a library that demonstrates this concretely which can be used out-of-the-box.
The include/ folder in our repository contains the header-only CUDA portion of our code, which can be integrated into frameworks relying on CUDA.
We also have a Pytorch module which can be installed with:

\begin{footnotesize}
\begin{lstlisting}[frame=single,rulecolor=\color{black}]
pip install -e .
\end{lstlisting}
\end{footnotesize}

Once installed, the module can be imported by putting `import ibp' in your Python file.

%\theendnotes

\end{document}